\documentclass{article}

 \usepackage[main, final]{neurips_2026}

\usepackage[utf8]{inputenc} 
\usepackage[T1]{fontenc}    
\usepackage{hyperref}       
\usepackage{url}            
\usepackage{booktabs}       
\usepackage{amsfonts}       
\usepackage{nicefrac}       
\usepackage{microtype}      
\usepackage{xcolor}         

\usepackage{amsmath}
\usepackage{amsthm}

\usepackage{algorithm}
\usepackage{algpseudocode}
\usepackage[T1]{fontenc}

\usepackage{multirow}
\usepackage{graphicx} 
\usepackage{subcaption}
\usepackage{colortbl}

\usepackage{titletoc}

\usepackage{enumitem}

\usepackage{adjustbox}

\definecolor{fbgreen}{RGB}{36, 53, 240}
\definecolor{struckgray}{RGB}{150, 150, 150}

\definecolor{easyBg}{HTML}{EAF3DE}
\definecolor{easyFg}{HTML}{3B6D11}
\definecolor{mediumBg}{HTML}{E6F1FB}
\definecolor{mediumFg}{HTML}{185FA5}
\definecolor{hardBg}{HTML}{FAEEDA}
\definecolor{hardFg}{HTML}{854F0B}
\definecolor{vhardBg}{HTML}{FCEBEB}
\definecolor{vhardFg}{HTML}{A32D2D}

\newcommand{\deasy}{\colorbox{easyBg}{\textcolor{easyFg}{\small\textbf{Easy}}}}
\newcommand{\dmedium}{\colorbox{mediumBg}{\textcolor{mediumFg}{\small\textbf{Medium}}}}
\newcommand{\dhard}{\colorbox{hardBg}{\textcolor{hardFg}{\small\textbf{Hard}}}}
\newcommand{\dvhard}{\colorbox{vhardBg}{\textcolor{vhardFg}{\small\textbf{Very Hard}}}}

\newcommand{\lead}[1]{\textbf{#1 \hspace{3pt}}}

\usepackage[most]{tcolorbox}
\definecolor{metablue}{RGB}{0, 82, 204}       
\definecolor{metabluebg}{RGB}{235, 242, 255}  

\newtcolorbox{bluebox}[1][]{
    enhanced,
    colback=metabluebg,   
    colframe=metablue, 
    boxrule=1.5pt,          
    arc=3pt,                
    auto outer arc,
    left=4pt, right=4pt, top=4pt, bottom=4pt, 
    drop shadow=black!10!white, 
    fonttitle=\bfseries\sffamily,
    attach boxed title to top left={xshift=10pt, yshift=-10pt},
    boxed title style={colback=blue!60!black, enhanced, frame hidden, arc=2pt}
}

\definecolor{baselinebg}{gray}{1}   
\definecolor{ours}{HTML}{378ADD}

\newcommand{\best}[1]{\textbf{#1}}
\newcommand{\second}[1]{\underline{#1}}

\DeclareMathOperator*{\E}{\mathbb{E}}
\DeclareMathOperator*{\argmin}{arg\,min}
\DeclareMathOperator*{\argmax}{arg\,max}

\usepackage{ulem} 

\title{Behavioral Foundation Models for Quality Diversity}

\author{%
  Nazim Bendib \\
  Sorbonne Université, ISIR\\
  Paris, France \\
  \texttt{bendib@isir.upmc.fr} \\
  \And
  Nicolas Perrin-Gilbert \\
  Sorbonne Université, ISIR\\
  Paris, France \\
  \texttt{nicolas.perrin@isir.upmc.fr} \\
  \And
  Olivier Sigaud \\
  Sorbonne Université, ISIR\\
  Paris, France \\
  \texttt{olivier.sigaud@isir.upmc.fr} \\
}

\begin{document}

\maketitle

\begin{abstract}

Behavioral Foundation Models (BFMs) are an emerging paradigm in reinforcement learning, playing a role analogous to large language models in natural language processing: they have shown remarkable versatility, enabling zero-shot performance, fast imitation, and online adaptation, all by exploiting the structure of a latent space.
In this work, we investigate whether the latent behavioral space induced by BFMs can serve as an effective search space to discover large repertoires of behaviorally diverse and high-performing policies through Quality-Diversity (QD) methods.
While QD methods generally search directly in high-dimensional policy parameter space, in this paper, we present BFM-QD, a framework that performs QD search in the compact latent space of a BFM. 
We further show that the BFM-QD framework provides a closed-form, gradient-free policy improvement operator that approximates a policy gradient update, but requires no critic training and no backpropagation. 
Across continuous-control benchmarks spanning dense locomotion, sparse navigation, and contact-rich manipulation, BFM-QD consistently outperforms parameter-space baselines, with particularly stark gains in sparse and deceptive settings, where  all tested parameter-space QD methods collapse to near-zero performance. 
These results show the effectiveness of the BFM-QD framework, benefiting from the synergy between dimensionality reduction of the search space and offline pretraining from diverse behavioral data. This positions BFMs as a general-purpose backbone for QD optimization, extending their utility beyond zero-shot task solving to the discovery of diverse behavioral repertoires.
\end{abstract}

\section{Introduction}

The history of machine learning is marked by a recent and decisive shift: rather than training models from scratch for each new task, the field has converged on the paradigm of large-scale pretraining followed by downstream reuse.
In natural language processing, this inflection point arrived with large language models: pretrained representations so rich that a single backbone could be adapted, prompted, or searched over to solve tasks that no individual fine-tuned model could reach alone. Reinforcement Learning (RL) appears to be approaching an analogous moment.

Two families are emerging as the RL analogue of this paradigm. 
On one hand, Vision-Language-Action models (VLAs) \cite{ma2024survey} benefit from large-scale pretraining on expert demonstrations mainly for text-conditioned robotic manipulation tasks. 
On the other hand,  Behavioral Foundation Models (BFMs) \cite{barreto2017successor,touati2021learning, tirinzoni2025zero} take a complementary approach.
Trained once on large offline non-expert datasets of diverse agent behavior, a BFM encodes a diverse behavioral manifold into a compact latent space $\mathcal{Z} \subset \mathbb{R}^d$, such that varying a single continuous vector $z$ is sufficient to instantiate a wide spectrum of motor skills, navigation strategies, and manipulation primitives, all without any additional training. This representational efficiency has enabled remarkable results in zero-shot task generalization \cite{touati2021learning, touati2022does}, fast imitation \cite{pirotta2023fast}, and online adaptation \cite{sikchi2025fast}, placing BFMs at the frontier of generalist agent design.
Yet a striking limitation unites all existing work on BFMs: they all extract exactly one optimal behavior from the model. Given a task, the latent code $z^*$ that maximizes expected return is inferred or optimized, and the resulting single policy is deployed. The vast behavioral structure embedded in its latent space, containing diverse, distinct, and potentially complementary behaviors, is underexploited.
From this perspective, this paper asks a simple but consequential question: rather than extracting a single optimal behavior from a BFM, can we discover a large repertoire of diverse, high-performing behaviors from its latent space?

Quality-Diversity (QD)~\cite{mouret2015illuminating} is the natural framework to unlock this diversity: it discovers large repertoires of diverse, high-performing behaviors. Such repertoires matter because a single policy is optimal only under fixed assumptions. An archive enables, for example, damage recovery (selecting a policy that avoids a damaged leg without retraining), sim-to-real transfer (testing several candidates on hardware), and skill libraries for hierarchical control.
Yet standard QD methods search directly in the policy parameter space $\Theta \subset \mathbb{R}^D$, which is high-dimensional, unstructured, and almost entirely composed of incoherent behaviors (random perturbations of network weights almost never produce rewarding behaviors). This is not a limitation of the QD objective, but of the search space it is applied to.
The key insight is that in a well-trained BFM, the parameter space $\mathcal{Z}$ is behaviorally rich: points $z \in \mathcal{Z}$ in the latent space decode to behaviors that cover a wide spectrum of task-relevant features. So, by design, $\mathcal{Z}$ is exactly the kind of search space QD needs.

We propose \textbf{BFM-QD}, a framework that exploits this synergy by relocating QD search from the policy parameter space to the latent space of a pretrained BFM, providing an ideal search space for QD without modifying its objective.
Across a benchmark spanning dense locomotion (Walker, HalfCheetah), long-horizon sparse navigation (AntMaze-medium), and contact-rich manipulation (Cube-single), BFM-QD consistently outperforms all parameter-space baselines. On easy and dense tasks, the margin is modest, but convergence is faster. On hard and sparse tasks, the gap is stark: all parameter-space methods fail, while BFM-QD maintains strong behavioral coverage and high performance.
Together, these results position BFMs not merely as tools for single-task policy extraction, but as general-purpose backbones for QD, extending their utility from zero-shot and few-shot task solving to the generation of large repertoires of efficient policies.

\textbf{Contributions:}
In this paper:
\begin{itemize}[leftmargin=0.7cm]
    \item We present BFM-QD, the first framework to use the latent space of a BFM as the search space for QD optimization.
    \item We show that BFMs provide a closed-form, gradient-free policy improvement operator that matches a policy-gradient variation operator, requiring no critic training and no backpropagation.
    \item Through an extensive benchmark spanning dense locomotion, sparse navigation, and contact-rich manipulation, we show that BFM-QD consistently outperforms all parameter-space baselines, with stark gains in sparse and deceptive settings where all tested parameter-space QD methods fail.
    
    \item We characterize two key properties of the BFM latent space: behavioral richness, where random latent vectors decode to diverse efficient behaviors, and a structure where interpolation between solutions produces meaningful offspring.

    \item We show that QD performance saturates early in BFM pretraining, well before zero-shot performance converges, revealing BFM-QD does not require a fully trained BFM.
    
\end{itemize}

\section{Related Work}

\subsection{Behavioral Foundation Models}
Behavioral Foundation Models are policies trained on large offline datasets of diverse behaviors, providing a pretrained latent space that can be adapted to new tasks without additional training. 
Successor Features \cite{dayan1993improving, barreto2017successor, borsa2018universal} provide an early instance of this paradigm, decomposing the value function into task-independent state features and task-specific reward weights, enabling zero-shot transfer across reward functions. 
Forward-Backward representations \cite{touati2021learning} extend this idea by jointly learning a forward map over successor measures and a backward map that projects reward functions into the latent space, enabling closed-form task inference and zero-shot policy retrieval. 
TD-JEPA \cite{bagatella2025td} further advances this direction by leveraging latent-predictive representations that capture long-term, policy-conditioned dynamics from offline reward-free data.
Beyond zero-shot task solving, BFMs have demonstrated broad utility as reusable behavioral backbones. They have been applied to fast imitation learning \cite{pirotta2023fast} using a small set of expert demonstrations, and to fast online adaptation \cite{sikchi2025fast}, where the latent space enables rapid fine-tuning to new tasks with minimal environment interaction. 
More recently, BFMs have been scaled to humanoid robot control \cite{tirinzoni2025zero}, demonstrating that the learned latent behavioral structure transfers to high-dimensional, contact-rich settings.
Our work extends this line of research in a new direction: rather than extracting a single optimal behavior from the BFM, we use its latent space as the search space for quality-diversity optimization, discovering entire repertoires of diverse and high-performing behaviors.

\subsection{Quality-Diversity algorithms}
Quality-Diversity algorithms aim to discover large collections of diverse, high-performing solutions rather than a single optimum. MAP-Elites (ME) \cite{mouret2015illuminating} is the canonical algorithm, maintaining an archive of elite solutions indexed by behavioral descriptors and evolving them through random mutations. CMA-ME \cite{fontaine2020covariance} improves upon this by replacing Gaussian mutation with Covariance Matrix Adaptation (CMA-ES) \cite{hansen2003reducing}, enabling more structured and efficient parameter-space search. However, these methods rely purely on undirected exploration and struggle in environments where random parameter perturbations rarely produce rewarding behaviors.
To address this, policy gradient methods have been integrated into QD. PGA-ME \cite{nilsson2021policy} augments ME with TD3 \cite{fujimoto2018addressing} actor updates, using a learned critic to guide mutations toward higher-fitness regions. QDRL \cite{lim2023understanding} similarly combines QD with deep RL, alternating between diversity-seeking and reward-maximizing updates. These approaches improve performance on dense-reward tasks but rely on a well-trained critic, which degrades under sparse or deceptive rewards.
A complementary line of work augments the gradient-based variation operator with explicit descriptor conditioning.
DCG-ME \cite{faldor2023map} enhances the policy gradient variation operator with a descriptor-conditioned critic, providing gradients that jointly optimize fitness and target descriptors. DCRL-ME \cite{faldor2025synergizing} builds on this by additionally exploiting the descriptor-conditioned actor as a generative model, injecting diverse solutions directly into the offspring batch at each generation.
Our work is complementary to these: rather than improving the descriptor or critic, we relocate the search space entirely to the latent space of a pretrained BFM.
Another line of work searches learned latent spaces. Latent space illumination~\cite{fontaine2021illuminating} runs QD in the latent space of a GAN to generate game levels; it is a content-generation task without policies or control. Policy Manifold Search~\cite{rakicevic2021policy} and DDE-Elites~\cite{gaier2020discovering} encode archive solutions with an autoencoder over policy parameters, so the latent space is learned online from the archive. In contrast, we search the latent space of a policy-generating model that is pretrained offline without rewards.

\section{Background}
\subsection{Behavioral Foundation Models}

A Behavioral Foundation Model (BFM) is an agent pretrained on a reward-free offline dataset of environment transitions, such that it can produce optimal policies for a large class of reward functions specified at test time, without any additional training or planning.
Most BFMs are built on top of the successor measure \cite{blier2021learning, dayan1993improving} framework. The successor measure of a policy $\pi$ at state-action $(s, a)$ is the discounted distribution of future states:
\begin{equation}
    M^\pi(X \mid s, a) := \sum_{t \geq 0} \gamma^t \Pr(s_t \in X \mid s_0=s, a_0=a, \pi), \quad \forall X \subset \mathcal{S}.
\end{equation}
Successor measures disentangle environment dynamics from the reward function: for any reward $r$ and policy $\pi$, the Q-function decomposes as $Q^\pi_r(s,a) = M^\pi r(s,a)$. Given a feature map $\phi: \mathcal{S} \rightarrow \mathbb{R}^d$, the corresponding successor features are $\psi^\pi(s,a) = \sum_{t \geq 0} \gamma^t \mathbb{E}[\phi(s_{t+1}) \mid s, a, \pi]$, which allow expressing the Q-function compactly as $Q^\pi_r(s,a) = \psi^\pi(s,a)^\top z$ for any reward $r_z(s) = \phi(s)^\top z$.

BFMs instantiate this framework by learning a family of policies $\{\pi_z\}_{z \in \mathcal{Z}}$ parameterized by a latent code $z \in \mathcal{Z} \subset \mathbb{R}^d$, alongside their successor features $\psi(s, a, z)$. At test time, given a reward function $r$, the optimal latent code is inferred by projecting $r$ onto the learned features:
\begin{equation}
    z_r =  \argmin_{z \in \mathbb{R}^d} \mathbb{E}_{s \sim \rho}\big[ ( r(s) - \phi(s)^\top z )^2 \big] = \mathbb{E}_{s \sim \rho}[\phi(s)\phi(s)^\top]^{-1} \mathbb{E}_{s \sim \rho}[\phi(s) r(s)],
    \label{eq:usf_inference}
\end{equation}
and the policy $\pi_{z_r}$ is deployed zero-shot. In practice $z_r$ is projected to a hypersphere of radius $\sqrt{d_z}$ with $d_z = \text{dim}(\mathcal{Z})$.

Different BFMs differ in how they learn $\phi$.
Forward-Backward (FB) representations \cite{touati2021learning} jointly learn $\phi$ and $\psi$ through a successor measure consistency loss, enabling closed-form task inference at test time. 
TD-JEPA \cite{bagatella2025td} instead learns $\phi$ and $\psi$ through a temporal-difference latent-predictive loss, training a policy-conditioned predictor that approximates successor features directly in latent space. 
Other BFMs are presented in Appendix~\ref{app:background_bfm}.
In all cases, the result is a single pretrained model $\pi_{\text{BFM}}(.|z)$ whose latent space $\mathcal{Z}$ encodes a wide range of behaviors, all accessible by varying $z$.

\subsection{Quality-Diversity algorithms}

Quality-Diversity (QD) algorithms aim to discover a large collection of diverse, high-performing solutions rather than converging to a single optimum. A QD problem is defined by a fitness function $r: \mathcal{S} \rightarrow \mathbb{R}$, measuring how good a solution is, and a behavioral descriptor $\beta: \mathcal{S} \rightarrow \mathcal{B}$, 
a low-dimensional summary of how a solution behaves (e.g., the foot contact pattern of a walking robot).
The descriptor space $\mathcal{B}$ is discretized into a finite grid of cells, and the goal is to find the best possible solution for every cell, that is, to fill the grid with a diverse repertoire of high-performing policies, each with a distinct behavior.
Solutions are maintained in an archive $\mathcal{C}$ storing at most one elite per cell; a candidate replaces the existing elite only if it achieves higher fitness in the same cell.

ME \cite{mouret2015illuminating} is the canonical QD algorithm: parents are sampled from the archive, perturbed by a variation operator, evaluated, and inserted if they improve their cell. The typical variation operator is a Gaussian mutation, but it can be replaced with more sophisticated methods, such as covariance-based adaptation \cite{fontaine2020covariance} or policy gradient updates that leverage a learned critic to guide mutations toward higher-fitness regions \cite{nilsson2021policy, lim2023understanding, faldor2023map, faldor2025synergizing}. See Appendix~\ref{app:qd_background} for more details.
Each generation thus repeats four steps: select parents from the archive, vary them, evaluate fitness and descriptor with a rollout, and insert offspring into their cells (Algorithm~\ref{alg:base_qd}).

The performance of QD methods is measured by three metrics: 
\textbf{coverage}, the fraction of cells populated in the archive; 
the \textbf{QD-score}, the sum of fitness values across all filled cells, which jointly captures whether the repertoire is both diverse and high-performing; and \textbf{maximum fitness}, the fitness of the single best solution found, measuring peak performance regardless of diversity.

\section{Method}

\begin{figure}
    \centering
    \includegraphics[width=1\linewidth]{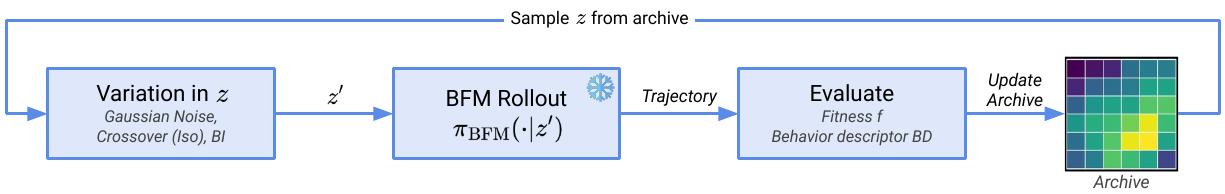}
    \caption{\textbf{Overview of BFM-QD.} A BFM is pretrained once on an offline dataset and then frozen. The loop is one QD generation: latent codes $z$ are sampled from the archive, mutated into $z'$, and evaluated by rolling out $\pi_{\text{BFM}}(\cdot|z')$. Offspring update the archive.}
    \label{fig:illustration}
    \vspace{-0.5cm}
\end{figure}

Standard QD methods parameterize policies as neural networks $\theta \in \Theta$ and mutate directly in $\Theta$. 
We instead leverage a pretrained BFM, specifically its pretrained actor $\pi_{\text{BFM}}: \mathcal{S} \times \mathcal{Z} \rightarrow \mathcal{A}$, which produces diverse behaviors by just varying $z \in \mathcal{Z}$. The BFM is trained once on an offline dataset and then kept frozen throughout all QD runs.

\subsection{Latent Space Quality-Diversity}

The central observation of our work is that $\mathcal{Z}$ is a far more efficient search space for QD than $\Theta$: it is low-dimensional ($d_z \ll |\theta|$) and behaviorally rich.
We therefore propose to run QD directly in $\mathcal{Z}$. The archive stores latent codes $z$ rather than policy weights $\theta$, and all variation operators act on $z$. Each candidate is evaluated by rolling out $\pi_{\text{BFM}}(\cdot \mid z)$ in the environment and computing its fitness and behavioral descriptor. The QD loop is otherwise standard: solutions compete for cells in the archive based on fitness, and the archive is used to seed the next generation of mutations.
Since the BFM is trained with a normalization constraint, we re-project every mutated candidate back onto the sphere after perturbation. See Figure~\ref{fig:illustration} for an overview. 
See Algorithm~\ref{pseudo:FB-QD} for pseudocode.


This constitutes our framework, which we refer to as \textbf{BFM-QD}. A key property of the BFM-QD framework is that it is modular by design, meaning any BFM can be paired with any QD optimizer, yielding variants such as FB-ME, FB-CMA-ME, TD-JEPA-ME, or TD-JEPA-CMA-ME.

\subsection{Closed-Form Policy Improvement via Backward Inference}
\label{method:bi}

Standard QD methods rely on gradient-free exploration, which is sample-inefficient. Recent work has addressed this by integrating policy gradient methods into QD~\cite{nilsson2021policy, faldor2025synergizing}: a critic is trained on the fitness function, and policies are updated to maximize it, improving solution quality while the archive maintains diversity. However, this requires training and maintaining a critic online, and doing multiple backpropagations to update the solutions.

BFMs enable sidestepping this limitation. A key advantage of BFMs is their zero-shot capabilities:
Given a fitness function and a set of evaluated transitions, the latent code $z^*$ corresponding to the optimal policy for the observed reward function can be estimated in closed form following Eq~\ref{eq:usf_inference}. This inference requires no gradient computation, no critic training, and no additional environment interactions beyond those already collected during archive evaluation.
We use this inference, which we call the \textbf{Backward Inference (BI)} operator, as a directed variation operator. For each solution $z$ selected from the archive, we then produce an offspring by interpolating toward this global $z^*$:
\begin{equation}
    z' \leftarrow (1-\alpha) \cdot z + \alpha \cdot z^*,
    \label{eq:bi_operator}
\end{equation}
where $\alpha \in ]0, 1[$ controls the step size toward the inferred optimum. $z'$ is then normalized and projected back onto the BFM latent sphere. 
Note that we do not set $z' \leftarrow z^*$ directly: doing so for every solution would collapse the entire archive toward a single behavior, destroying the diversity that QD is designed to maintain. Instead, we push each solution individually toward higher fitness while preserving its distinctiveness.
At each generation, we (i) sample state--reward pairs $(s, r(s))$ from the replay buffer, which stores all past evaluation trajectories; (ii) compute $z^*$ with Eq.~\ref{eq:usf_inference}; and (iii) for each parent $z$ selected from the archive, produce an offspring with Eq.~\ref{eq:bi_operator} and project it onto the sphere. 
Following PGA-ME, half of the offspring are produced by BI ($\alpha=0.02$) and half by Gaussian mutation ($\sigma=1$).
A pseudo-algorithm is presented in Algorithm~\ref{pseudo:FB-BI}.

Interestingly, this interpolation is a Newton step on a local quadratic surrogate of the policy improvement objective, moving $z$ toward $z^*$ (Appendix~\ref{app:proposition}). The step itself is generic; the BFM makes it practical by providing $z^*$ in closed form, with no critic training and no backpropagation. The approximation error is proportional to the reward projection residual.

\section{Experiments}
\label{section:experiments}
We evaluate BFM-QD on a suite of continuous control tasks spanning a spectrum of reward densities, from dense locomotion to sparse navigation and contact-rich manipulation. Our experiments are designed to answer four questions:
\textbf{(1)} Does BFM-QD outperform standard QD baselines across task types and reward densities?
\textbf{(2)} What properties of the FB latent space drive BFM-QD's performance?
\textbf{(3)} Can Backward Inference in BFMs replace policy gradient updates?
\textbf{(4)} Does BFM-QD require a large pretraining budget to be effective?
 
\subsection{Experimental Setup}

\lead{Environments.} 
We evaluate on 18 tasks spanning four environments and a spectrum of difficulty levels. 
\textbf{Walker2D\_uni} and \textbf{HalfCheetah\_uni} are standard continuous control benchmarks from QDax \cite{chalumeau2024qdax}, representing dense, simple locomotion, with pretraining data collected via Random Network Distillation (RND) \cite{burda2018exploration}.
\textbf{AntMaze-medium} from OGBench \cite{park2024ogbench} uses the \texttt{antmaze-medium-explore} dataset collected by random policies and introduces long-horizon navigation with both dense and sparse rewards. 
\textbf{Cube-single} from OGBench uses the \texttt{cube-single-noisy} dataset, consisting of noisy trajectories, and introduces manipulation under sparse rewards (hardest setting). 
Full details of environments, behavioral descriptors, and fitness functions are provided in Appendix~\ref{appendix:envs}.

\lead{Baselines.}
We compare BFM-QD against five parameter-space baselines.
\textbf{ME} \cite{mouret2015illuminating} is the standard QD algorithm, operating in MLP parameter space with Gaussian mutation.
\textbf{CMA-ME} \cite{fontaine2020covariance} augments ME with CMA-ES.
\textbf{DCRL-ME} \cite{faldor2025synergizing} combines ME with TD3 policy gradient using a descriptor-conditioned policy.
\textbf{DDE-Elites} \cite{gaier2020discovering} iteratively trains a Variational Auto-Encoder (VAE) \cite{pinheiro2021variational} on the archive solutions to learn a compact latent space for more sample-efficient search. This baseline \emph{isolates the effect of searching in a low-dimensional space.}
\textbf{ME+Pretrain} initializes MLP policies via offline TD3 trained on the same dataset as the BFM, using the task fitness function as a reward. This baseline \emph{isolates the effect of pretraining on diverse data}. All baselines use the same MLP architecture for a fair comparison. For more details, see Appendix~\ref{appendix:baseline_details}.

\lead{BFM-QD Variants.}
We instantiate three BFM-QD variants:
\textbf{FB-ME} pairs FB with ME using Gaussian mutation;
\textbf{FB-CMA-ME} pairs FB with CMA-ME;
\textbf{TD-JEPA-CMA-ME} pairs TD-JEPA with CMA-ME.
Other variants of BFM-QD are presented in Appendix~\ref{app:bfm_comparison}.

\lead{Implementation Details.}
The FB model is pretrained once per environment on the offline dataset and kept frozen throughout all QD runs (See Appendix~\ref{app:compute} for reported compute time).
The latent space has dimension $d_z = 50$. The archive is a $50 \times 50$ grid. 
Each run consists of $500$ generations of $400$ parallel rollouts, for $200$k evaluations in total. A rollout is one full episode of $500$ steps (Walker, HalfCheetah) or $1000$ steps (AntMaze, Cube-single), so a run collects $100$M or $200$M transitions, respectively ($200$k or $400$k per generation).
Since each environment contains multiple tasks with different reward scales, we normalize QD-scores using min-max normalization before aggregating across tasks (see Appendix~\ref{appendix:envs} for details).
For parameter-based methods, we use MLPs of the shape (state\_dim, 128, 128, action\_dim).
All results are averaged over 5 random seeds unless stated otherwise. Standard deviations are reported.

\subsection{Main Results}

\begin{figure}
    \centering
    \includegraphics[width=0.9\linewidth]{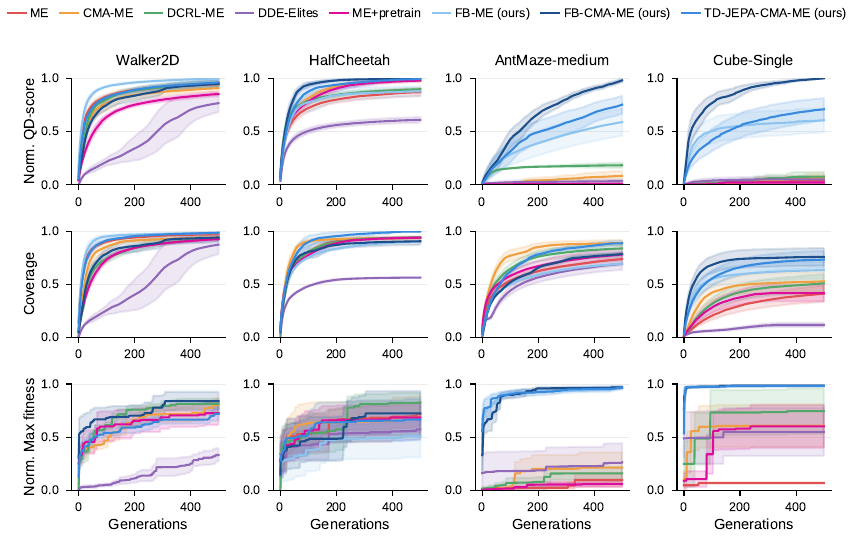}
    \caption{\textbf{Normalized QD-score, coverage, and normalized max fitness across all environments.} BFM-QD variants consistently match or outperform all baselines, especially on harder tasks.}
    \label{fig:main}
    \vspace{-0.5cm}
\end{figure}

Figure~\ref{fig:main} reports the QD-score, coverage, and maximum fitness across all methods and environments. BFM-QD variants consistently achieve the highest QD-score across all tasks. On dense locomotion tasks (Walker, HalfCheetah), the margin over parameter-space baselines is modest, but BFM-QD variants converge faster, confirming that the behavioral prior accelerates search even when the task is easy. Notably, BFM-QD variants achieve this despite operating in a far lower-dimensional space ($d_z = 50$ versus ${\sim}20k$ parameters), demonstrating that the BFM latent space is sufficiently expressive to reproduce near-optimal behaviors across a wide range of fitness functions.

The advantage of BFM-QD grows substantially as task difficulty increases. On AntMaze-medium and Cube-single, all parameter-space baselines collapse to near-zero QD-score and low coverage, while BFM-QD variants maintain high performance across all metrics. Crucially, BFM-QD variants also achieve substantially higher maximum fitness on hard tasks. 
Two additional comparisons sharpen this finding. \textbf{DDE-Elites}, which also searches in a compact latent space, fails to match the BFM-QD performance, demonstrating that dimensionality reduction alone is not sufficient: what matters is the behavioral richness of the latent space, not its dimensionality. Similarly, \textbf{ME+Pretrain}, which benefits from offline pretraining, also fails, demonstrating that the benefit of BFM-QD is not simply a matter of pretraining on diverse data, but of searching in the structured latent space the BFM provides. 
Two further baselines that combine offline pretraining with latent-space search (DDE-Elites+Pretrain with TD3 or BC policies) and Policy Manifold Search also fall short of FB-ME (Appendix~\ref{app:pretrain_latent}).

\begin{bluebox}
\textbf{Takeaway:} 
Searching in the BFM latent space consistently outperforms searching in the parameter space, with stark gains in sparse and deceptive settings.  
The gains do not come from dimensionality reduction or pretraining alone, 
but from the compact behavioral representation of the BFM.
\end{bluebox}

\subsection{Expressivity of the latent space}

We investigate two properties of the FB latent space that underpin the success of BFM-QD: the expressivity of the learned latent space and its smooth structure, where linear interpolations between solutions produce behaviorally intermediate policies.

\begin{figure}
    \centering
    \begin{subfigure}[b]{0.41\linewidth}
        \centering
        \includegraphics[width=\linewidth]{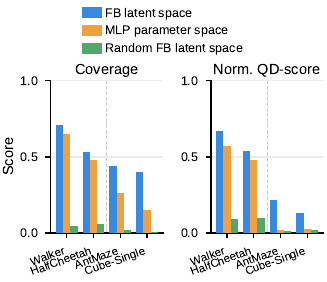}
        \caption{\textbf{Random sampling.} Coverage and QD-score of 10,000 random 
        FB latent samples vs. random MLP samples.}
        \label{fig:random_sampling}
    \end{subfigure}
    \hspace{3pt}
    \begin{subfigure}[b]{0.50\linewidth}
        \centering
        \includegraphics[width=\linewidth]{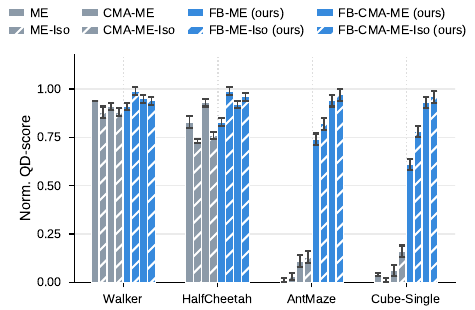}
        \caption{\textbf{Interpolation operator.} QD-score for ME and CMA-ME, with and without the isoline operator, in parameter space and FB latent space.}
        \label{fig:interpolation}
    \end{subfigure}
    \caption{\textbf{The FB latent space is expressive and behaviorally rich, and supports interpolation between behaviors.} (a) Random FB latent codes yield higher coverage and QD-score compared to random MLP policies. (b) Interpolation between solutions is only beneficial in the FB latent space, not in parameter space.}
    \label{fig:expressivity}
    \vspace{-0.5cm}
\end{figure}

\subsubsection{The latent space is behaviorally rich by construction}
\label{sec:expressivity}

To isolate the contribution of the behavioral prior from the QD optimizer, we remove all optimization entirely and evaluate the quality of random sampling alone.

\lead{Experiment}
We sample $N=10{,}000$ random latent vectors from 
\textbf{(1)} a trained FB latent space, 
\textbf{(2)} the MLP parameter space, 
and \textbf{(3)} an untrained FB latent space (same architecture, random weights) to further isolate the contribution of pretraining from that of architecture.
We report the resulting coverage and QD-score for each environment. No optimization is performed in any condition.

\lead{Results}
Results are shown in Figure~\ref{fig:random_sampling}. Even without optimization, random FB samples already achieve higher coverage and QD-score than random MLP samples across all environments, with the gap growing monotonically with task difficulty (AntMaze-medium and Cube-single), where random MLP samples produce near-zero QD-score (see Appendix~\ref{appendix:random_sampling} for archive visualization). 
Crucially, the untrained FB model performs significantly worse than the trained FB, confirming that this behavioral richness stems specifically from offline pretraining, not from the architecture or the geometry of the latent sphere.

\begin{bluebox}
\lead{Takeaway:}
Through pretraining, the FB latent space naturally produces behaviors that cover a wide spectrum of task-relevant features.
\end{bluebox}

\subsubsection{The latent space structure makes interpolation meaningful}
\label{exp:interpolation}

A key property of the BFM latent space is its geometric smoothness: nearby latent codes induce behaviorally similar policies. We investigate whether this smoothness is sufficient to make interpolation between archive solutions semantically meaningful, enabling a class of efficient crossover-based variation operators. By contrast, such operators can be destructive in parameter space as the interpolant typically collapses to an incoherent solution \cite{entezari2021role}.
We introduce an isoline operator (Iso) that, given two parent solutions $x_1, x_2$ from the archive, produces an offspring by linear interpolation:
\[
x_{\text{offspring}} = (1-\alpha)\, x_1 + \alpha\, x_2, \quad \alpha \sim \mathcal{U}[0, 1].
\]
A pseudo-algorithm is presented in Algorithm~\ref{pseudo:FB-Iso}.

\lead{Experiment}
We compare ME and CMA-ME with and without the interpolation operator, both in parameter space (ME-Iso, CMA-ME-Iso) and the FB latent space (FB-ME-Iso, FB-CMA-ME-Iso), isolating whether the gain comes from the operator itself or the smoothness of the latent space.

\lead{Results}
Results are shown in Figure~\ref{fig:interpolation}. Adding the isoline operator to parameter-space methods (ME-Iso, CMA-ME-Iso) yields no consistent improvement, confirming that interpolation in parameter space can be destructive. In contrast, FB-ME-Iso and FB-CMA-ME-Iso consistently improve over their non-Iso counterparts, especially on AntMaze-medium and Cube-single.
This extra gain can be attributed to the structure of the latent space: interpolated codes produce behaviorally intermediate policies, enabling a better exploration of the descriptor space that random mutation cannot replicate.

\begin{bluebox}
\lead{Takeaway:}
Interpolation between solutions is only beneficial in the BFM latent space due to its structure and smoothness, unlike in the neural-network parameter space.
\end{bluebox}

\subsection{Can Backward Inference in BFMs replace policy gradient updates?}

\begin{figure}
    \centering
    \includegraphics[width=0.7\linewidth]{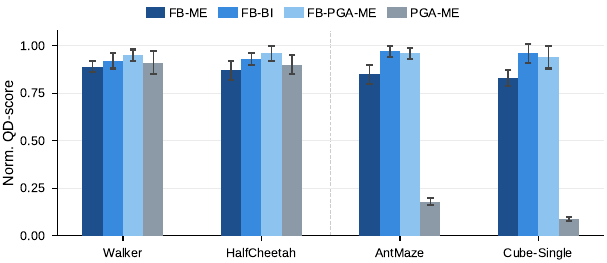}
    \caption{\textbf{The BFM Backward Inference (BI) matches a policy gradient update.} FB-BI (backward inference) matches FB-PGA-ME across all environments, while PGA-ME fails on hard tasks.}
    \label{fig:bi_operator}
    \vspace{-0.5cm}
\end{figure}

A natural question is whether gradient-based variation operators can further improve BFM-QD. PGA-ME~\cite{nilsson2021policy} augments ME with TD3 policy gradient updates, using a learned critic to guide mutations toward higher-fitness regions. However, BFMs and typically FB provide a closed-form estimate of the task vector that best explains an observed reward signal (Equation~\ref{eq:usf_inference}), requiring no critic training and no additional environment interactions. We investigate whether this zero-shot inference approach can substitute for a full policy gradient update.

\lead{Experiment}
We compare three methods: \textbf{PGA-ME}, which runs TD3 policy gradient updates in parameter space; \textbf{FB-PGA-ME}, which runs the same gradient updates but in the FB latent space; and \textbf{FB-BI}, which replaces gradient updates entirely with the Backward Inference (BI) operator (Equation~\ref{eq:bi_operator}). See Appendix~\ref{appendix:baseline_details} for more details.

\lead{Results}
Results are shown in Figure~\ref{fig:bi_operator}. PGA-ME matches FB-PGA-ME and FB-BI methods on locomotion tasks but drastically fails on AntMaze-medium and Cube-single, where the reward signal is too deceptive and sparse to train an accurate critic. FB-BI matches FB-PGA-ME across all environments, achieving the same QD-score without any critic training. 
This shows that closed-form backward inference can substitute for a learned critic.

\begin{bluebox}
\lead{Takeaway:}
The BFM Backward Inference (BI) operator matches a policy-gradient variation operator, without critic training or backpropagation.
\end{bluebox}

\subsection{BFM-QD does not require a large pretraining budget}
\label{training_budget}

We investigate whether BFM-QD requires a large BFM pretraining budget to be effective for QD search.
A priori, one might expect that QD performance follows zero-shot performance and thus requires a fully trained BFM. We test whether this holds by measuring QD performance and zero-shot performance at different stages of BFM training.

\lead{Experiment}
We train FB and extract checkpoints every 40k steps. At each checkpoint, we run FB-ME and report the resulting QD-score and the zero-shot performance of the checkpoint on the same task. We do this for (\texttt{energy efficiency |ant xy}) from AntMaze-medium and (\texttt{energy efficiency |cube xz}) from Cube-single.

\begin{figure}
    \centering
    \includegraphics[width=0.8\linewidth]{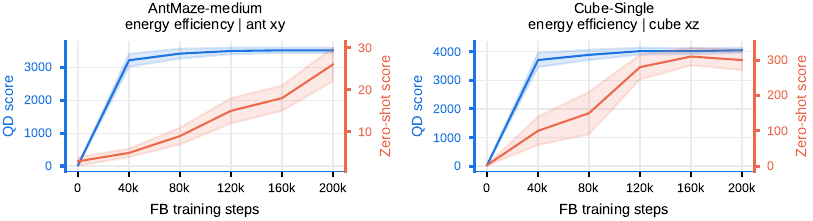}
    \caption{\textbf{BFM-QD does not require a fully trained BFM.} QD-score (blue) and zero-shot performance (red) as a function of FB training budget on AntMaze-medium and Cube-single}
    \label{fig:training_budget}
    \vspace{-0.5cm}
\end{figure}

\lead{Results}
Results are shown in Figure~\ref{fig:training_budget}. QD performance rises sharply in the first 40k training steps and plateaus early, well before zero-shot performance converges. In both environments, a partially trained BFM already enables strong QD performance, while zero-shot performance continues to improve long after the QD-score has saturated. This decoupling confirms that the latent space acquires sufficient behavioral structure for QD search early in training, even before it can reliably solve individual tasks zero-shot.

We distinguish two properties of the BFM latent space: \emph{behavioral diversity}, the repertoire of distinct behaviors it contains, and \emph{value precision}, the accuracy of $(\phi,\psi)$ for computing $z_r$ in closed form (Equation~\ref{eq:usf_inference}). Zero-shot solving requires both: diversity provides good candidates, and precision selects the best one. QD search requires only diversity. We attribute the decoupling in Figure~\ref{fig:training_budget} to the two properties emerging at different times: diversity is acquired early in pretraining, while value precision keeps improving later.

\begin{bluebox}
\lead{Takeaway:}
A partially trained BFM already provides a latent space rich enough for strong QD performance, amortizing pretraining cost further.
\end{bluebox}

\subsection{Additional Analyses}

Beyond the main experiments, we conduct seven additional analyses in Appendix~\ref{app:additional_results}, whose findings we summarize here:
\begin{enumerate}[leftmargin=0.7cm]

    \item \textbf{The choice of a BFM matters at scale (\ref{app:bfm_comparison}):} Across five BFMs (Laplacian~\cite{wu2018laplacian}, BYOL$\gamma$~\cite{lawson2025self}, FB, BTD-FB~\cite{bendib2026improving}, and TD-JEPA), zero-shot performance is a reliable proxy for QD performance; the gap between strong and weak BFMs widens sharply on hard tasks, making BFM quality a critical design decision in sparse settings, while weaker BFMs can still perform well on easy tasks.

    \item \textbf{Diversity aids optimization, not just coverage (\ref{app:diversity}):} FB-CMA-ME outperforms a single-objective CMA-ES \cite{hansen2016cma} in the same latent space even on maximum fitness, demonstrating that maintaining a diverse population acts as an implicit exploration bonus that prevents premature convergence to local optima in the latent space.
    
    \item \textbf{Exploration data dominates dataset quality (\ref{app:dataset}):} Random exploration datasets consistently outperform expert datasets for BFM pretraining, even when spatially restricted, because limited coverage in the behavioral space in the offline data results in bounded latent space diversity.

    \item \textbf{Pretraining and latent-space search are not sufficient (\ref{app:pretrain_latent}):} DDE-Elites initialized with policies pretrained offline on the BFM dataset (TD3 or behavioral cloning) and Policy Manifold Search do not match FB-ME, showing that the gains come from the structure of the BFM latent space and not from combining pretraining with a learned latent space.
    
    \item \textbf{Backward Inference outperforms sampling around $z^*$ (\ref{app:around}):} Sampling offspring around the inferred optimum confines the search to a small region of the latent space, whereas BI moves each parent individually toward $z^*$ and preserves diversity.

    \item \textbf{QD-collected data is a worse pretraining source than RND (\ref{app:qddata}):} A BFM pretrained on data collected by a QD search performs worse than one pretrained on RND data, because QD refines solutions near existing elites instead of covering new transitions.

    \item \textbf{Backward Inference is robust to its step size (\ref{app:alpha}):} FB-BI is stable for $\alpha\in[0.005,0.1]$ and degrades for larger steps, which pull offspring toward the single point $z^*$ and collapse diversity.

\end{enumerate}

\section{Limitations}
\label{section:limitation}

BFM-QD requires a behaviorally diverse offline dataset and a sufficiently pretrained BFM before any QD search can begin, though Section~\ref{training_budget} shows that full convergence of the BFM is not necessary.
Besides, a single BFM can be reused across all subsequent tasks and QD runs in the same environment, amortizing this cost (see Appendix~\ref{app:compute} for training time).
Furthermore, performance degrades with narrower datasets (Appendix~\ref{app:dataset}) and with data collected by QD instead of RND (Appendix~\ref{app:qddata}), as the diversity of the latent space is bounded by the diversity of the training data. 
A deeper limitation is that the frozen BFM imposes a hard ceiling on behavioral expressivity: novel behaviors outside the pretrained latent space remain inaccessible. Finetuning the BFM during the QD run is a promising direction we leave to future work.
Moreover, each BFM is trained for a single environment, so environment-agnostic QD remains open. Finally, as in standard QD, we assume the reward and behavioral descriptor are given; BFM-QD can only produce behaviors captured by the learned features $\phi(s)$, and we do not measure how far the discovered behaviors lie outside the pretraining data.

\section{Conclusion}

We proposed BFM-QD, a framework that replaces parameter-space exploration in QD search with exploration in the latent space of a pretrained BFM. BFM-QD sidesteps the fundamental bottleneck of standard QD methods by searching in a compact, behaviorally dense latent space, consistently outperforming parameter-space baselines across dense locomotion, sparse navigation, and contact-rich manipulation, with stark gains in sparse and deceptive settings where all tested parameter-space methods collapse. These results establish BFM latent space exploration as a principled foundation for the next generation of QD algorithms,  and position BFMs as general-purpose backbones for diversity-seeking optimization beyond their original use case of single-task policy extraction.

\section*{Acknowledgments}
Experiments presented in this paper were carried out using the HPC resources of IDRIS under the allocation 2025-[AD011016374] made by GENCI.
This work was supported by the Sorbonne Center for Artificial Intelligence (SCAI)

\textbf{Competing interests.} We declare no competing interests.

\bibliography{neurips_2026}
\bibliographystyle{plain}

\newpage
\appendix

\section*{Appendix Table of Contents}
\startcontents[appendix]
\printcontents[appendix]{}{0}{}

\newpage
\section{Proof of the Backward Inference Operator}
\label{app:proposition}

In the following, we denote $\pi_z(\cdot) := \pi_{\mathrm{BFM}}(\cdot \mid z)$ the policy instantiated by the BFM at latent code $z \in \mathcal{Z}$.Consider the policy improvement objective:
\begin{equation}
    J(z) = \mathbb{E}_{s \sim \rho}\bigl[-Q^{\pi_z}(s, \pi_z(s))\bigr],
\end{equation}
where $Q^{\pi_z}(s,a) = \mathbb{E}\bigl[\sum_{t \geq 0} \gamma^t r(s_t) \mid s_0 = s, a_0 = a, \pi_z\bigr]$ is the action-value function of policy $\pi_z$ under reward $r$, and $\rho$ is the mixture state distribution induced by the archive policies (equivalent to a replay buffer). 
A standard first-order policy gradient step on $J$ takes the form:
\begin{equation}
    z \leftarrow z - \alpha \, \nabla_z J(z).
\end{equation}
We will show that the BI operator (Equation~\ref{eq:bi_operator}) actually approximates a \emph{second-order Newton step} of $J$, which subsumes this update.

By construction of $z^*$ (Equation~\ref{eq:usf_inference}), the task vector $z^*$ is the least-squares projection of $r$ onto the feature space of $\phi$:
\begin{equation}
    z^* = \argmin_{z \in \mathbb{R}^{d_z}} \, \mathbb{E}_\rho\bigl[(r(s) - \phi(s)^\top z)^2\bigr],
\end{equation}
so that $r(s) \approx \phi(s)^\top z^*$ for all $s$ \footnote{This approximation holds exactly when $r$ lies in the linear span of $\phi$, and approximately otherwise, with error proportional to the projection residual $\|r - \phi^\top z^*\|_\rho$.}. Substituting into the definition of $Q^{\pi_z}$:
\begin{align}
    Q^{\pi_z}(s, a) 
    &= \mathbb{E}\!\left[\sum_{t \geq 0} \gamma^t r(s_t) \,\Big|\, s, a, \pi_z\right] \\
    &\approx \mathbb{E}\!\left[\sum_{t \geq 0} \gamma^t \phi(s_t)^\top z^* \,\Big|\, s, a, \pi_z\right] \\
    &= \underbrace{\mathbb{E}\!\left[\sum_{t \geq 0} \gamma^t \phi(s_t) \,\Big|\, s, a, \pi_z\right]}_{\psi^{\pi_z}(s,a)} \!\!\!{}^\top z^* 
    \;=\; \psi^{\pi_z}(s, a)^\top z^* 
    \;=\; Q^{\pi_z}_{z^*}(s, a),
\end{align}
where $\psi^{\pi_z}(s,a)$ are the successor features of $\pi_z$ \cite{barreto2017successor}. The policy gradient objective therefore becomes:
\begin{equation}
    J(z) \approx \mathbb{E}_{s \sim \rho}\bigl[-Q^{\pi_z}_{z^*}(s, \pi_z(s))\bigr]
    = \mathbb{E}_{s \sim \rho}\bigl[-\psi^{\pi_z}(s, \pi_z(s))^\top z^*\bigr].
\end{equation}
Denote this surrogate objective $\tilde{J}(z) := \mathbb{E}_{s \sim \rho}\bigl[-\psi^{\pi_z}(s, \pi_z(s))^\top z^*\bigr]$, so that $J(z) \approx \tilde{J}(z)$ with error proportional to the reward projection residual $\|r - \phi^\top z^*\|_2$.
By definition, $\pi_{z^*}$ is the greedy policy with respect to $Q^{\pi_{z^*}}_{z^*}$:
\begin{equation}
    \pi_{z^*}(s) = \argmax_a \, \psi^{\pi_{z^*}}(s,a)^\top z^*,
\end{equation}
and is therefore optimal for task $z^*$. 
By Bellman optimality~\cite{sutton1998reinforcement}, for any other policy $\pi_z$:
\begin{equation}
    Q^{\pi_{z^*}}_{z^*}(s, a) \geq Q^{\pi_z}_{z^*}(s, a) \quad \forall\, z,\, s,\, a,
\end{equation}
which implies $z^* = \argmin_z \tilde{J}(z)$ and therefore $\nabla_z \tilde{J}\big|_{z = z^*} = 0$.

A second-order Taylor expansion of $\tilde{J}$ around $z^*$ gives:
\begin{equation}
    \tilde{J}(z) = \tilde{J}(z^*) + \underbrace{\nabla_z \tilde{J}\big|_{z^*}^\top}_{=\,0}(z - z^*) 
    + \frac{1}{2}(z - z^*)^\top H\,(z - z^*) + O\!\left(\|z - z^*\|^3\right),
\end{equation}
where $H = \nabla^2_z \tilde{J}\big|_{z^*} \succeq 0$. Differentiating gives $\nabla_z \tilde{J}(z) \approx H(z - z^*)$. Assuming a strict positive definiteness of the hessian ($H \succ 0$), a Newton step on $\tilde{J}$ yields:
\begin{align}
    z &\leftarrow z - \alpha\, H^{-1} \nabla_z \tilde{J}(z) \\
      &= z - \alpha\, H^{-1} H(z - z^*) \\
      &= (1 - \alpha)\,z + \alpha\, z^*.
\end{align}

Projecting back onto the latent sphere $\|z\| = \sqrt{d_z}$ recovers the BI operator (Equation~\ref{eq:bi_operator}). 
A Newton step on $\tilde{J}$ therefore approximates a Newton step on the true objective $J$, with residual error proportional to the reward projection quality $\|r - \phi^\top z^*\|_2$.

\textbf{Remark.} 
The BI operator constitutes a \emph{second-order} policy gradient update in the BFM latent space. The critic gradient direction of PGA-ME is replaced by the analytically inferred $z^*$, obtained in closed form from archive rollouts with no critic training and no backpropagation. The proof extends to transition-dependent rewards $r(s, a)$ by replacing $\phi(s)$ with $\phi(s, a)$ throughout.

\newpage
\section{Background on Behavioral Foundation Models}
\label{app:background_bfm}

Behavioral foundation models~(BFMs) aim to pretrain a single agent from reward-free interactions such that it can produce optimal policies for any downstream reward at test time, with no or minimal additional learning. The central challenge is to learn a compact representation of the environment's long-term dynamics that is simultaneously expressive enough to cover all tasks and structured enough to support efficient zero-shot policy extraction.

\subsection{Successor Features}
\label{app:sf}

The standard \(Q\)-function conflates environment dynamics and task rewards, making it expensive to reuse across tasks. \cite{barreto2017successor} propose Successor Features~(SF) to decouple them. They assume that the reward on each transition can be linearly decomposed as
\begin{equation}
  r(s, a, s') \;=\; \phi(s, a, s')^{\!\top} \mathbf{w},
  \label{eq:sf_reward}
\end{equation}
where \(\phi : \mathcal{S} \times \mathcal{A} \times \mathcal{S} \to \mathbb{R}^d\) is a feature map shared across all tasks and \(\mathbf{w} \in \mathbb{R}^d\) is a task-specific reward weight vector. Under this factorization, the action-value function of any policy \(\pi\) on task \(\mathbf{w}\) decomposes as
\begin{equation}
  Q^{\pi}_{\mathbf{w}}(s, a)
  \;=\; \underbrace{\mathbb{E}\!\left[\sum_{t=0}^{\infty} \gamma^t
        \phi(s_t, a_t, s_{t+1}) \;\Big|\; s_0=s,\, a_0=a,\, \pi\right]}_{\psi^\pi(s,a)}
  {}^{\!\top} \mathbf{w}
  \;=\; \psi^\pi(s, a)^{\!\top} \mathbf{w},
  \label{eq:sf_decomp}
\end{equation}
where \(\psi^\pi(s, a) \in \mathbb{R}^d\) are the successor features of \(\pi\).

\paragraph{Training.}
Successor features satisfy a Bellman equation and can therefore be estimated via temporal difference~(TD) learning \cite{dayan1993improving}. Given a replay buffer \(\mathcal{D} = \{(s, a, s')\}\) and a fixed target policy \(\bar{\pi}\), the standard TD loss is
\begin{equation}
  \mathcal{L}_{\mathrm{SF}}(\psi)
  \;=\; \mathbb{E}_{(s,a,s') \sim \mathcal{D}}\!
        \left\| \psi(s, a) - \phi(s, a, s') - \gamma\, \bar{\psi}(s', \bar{\pi}(s'))
        \right\|^2,
  \label{eq:sf_loss}
\end{equation}
where \(\bar{\psi}\) denotes a stop-gradient (target network) copy of \(\psi\). Once \(\psi^\pi\) has been learned under any exploration policy, adapting to a new task \(\mathbf{w}'\) only requires re-fitting the scalar weights using linear regression on a handful of evaluated samples:
\[
    \mathbf{w}^* = \argmin_{\mathbf{w} \in \mathbb{R}^d} \mathbb{E}_{s \in \mathcal{D}} \big[ (\phi(s)^\top \mathbf{w} - r(s))^2 \big],
\]
and the zero-shot policy is \(\pi_{\mathbf{w}^*}(s) = \arg\max_a \psi(s, a, \mathbf{w}^*)^{\!\top} \mathbf{w}^*\), requiring no additional environment interaction.

The central limitation is that the feature map \(\phi\) must be specified in advance (hand-coded or learned), and the expressivity of SF is bounded by the linear span of \(\phi\), making the quality of the representation a critical bottleneck.
Some prominent approaches to learn \(\phi\) directly from data are described below:

\begin{itemize}
    \item \textbf{Autoencoder} \cite{chen2023auto} learns a compact representation $\phi(s)$ by minimizing the reconstruction error of the input state through an encoder-decoder architecture, capturing the most salient features of the state space. Here, $\phi$ is the encoder $\phi(s)$ and $f$ is the decoder:
    \[
    \min_{f,\phi} \mathbb{E}_{s \sim \mathcal{D}} \left[ (f(\phi(s)) - s)^2 \right].
    \]

    \item \textbf{Transition Model} learns representations by training a model to predict the next state $s_{t+1}$ given the current state-action pair $(s_t, a_t)$, thereby encoding the environment's local dynamics into the latent space. Here, $\phi$ is the encoder $\phi(s)$ and $f$ is the latent transition predictor:
    \[
    \min_{f,\phi} \mathbb{E}_{(s_t, a_t, s_{t+1}) \sim \mathcal{D}} \left[ (f(\phi(s_t), a_t) - s_{t+1})^2 \right].
    \]

    \item \textbf{Laplacian (Lap)} \cite{wu2018laplacian}
    learns representations by approximating the eigenfunctions of the graph Laplacian induced by an exploratory policy, encouraging temporally adjacent states to have similar representations while keeping all state representations globally spread apart via an orthonormality regularization. Here, $f$ is the state encoder:

    \[
    \min_{f}  \E_{(s_t, s_{t+1}) \sim \mathcal{D}} \left[ \| f(s_t) - f(s_{t+1}) \|^2 \right] +
    \mathbb{E}_{\substack{s\sim \mathcal{D} \\ s'\sim \mathcal{D}}} 
    \left[  (f(s)^\top f(s'))^2 - \|f(s)\|^2 - \|f(s')\|^2  \right].
    \]
    
    \item \textbf{Lower-Rank Approximation of the transition probability} \cite{touati2022does} factorizes the one-step transition probability $P(s'|s, a)$ into a low-rank product $f(s, a)^\top \phi(s')$, effectively performing a spectral decomposition of the transition operator. Here, $f$ and $\phi$ represent the left and right singular vectors of the transition matrix:
    \[
    \min_{f,\phi} \frac{1}{2} \mathbb{E}_{\substack{(s_t, a_t) \sim \mathcal{D} \\ s' \sim \mathcal{D}}} \left[ \left( (f(s_t, a_t)^\top \phi(s')) \right)^2 \right] - \mathbb{E}_{(s_t, a_t, s_{t+1}) \sim \mathcal{D}} \left[ f(s_t, a_t)^\top \phi(s_{t+1}) \right].
    \]

    \item \textbf{Lower-Rank Approximation of the successor measure} \cite{touati2022does} learns a low-rank factorization of the successor measure $M(s, s')$ using temporal difference learning. Here, $f$ and $\phi$ are the learned factors that represent the state and its temporally extended features:
    \[
    \min_{f,\phi} \mathbb{E}_{\substack{(s_t, s_{t+1}) \sim \mathcal{D} \\ s' \sim \mathcal{D}}} \left[ \left( f(s_t)^\top \phi(s') - \gamma f(s_{t+1})^\top \bar{\phi}(s') \right)^2 \right] - 2 \mathbb{E}_{(s_t, s_{t+1}) \sim \mathcal{D}} \left[ f(s_t)^\top \phi(s_{t+1}) \right].
    \]

    \item \textbf{Bootstrap Your Own Latent (BYOL) \cite{grill2020bootstrap}} \label{appendix:bfms:byol} learns a latent space by predicting the next state representation from the current state-action pair using a predictor and a target network. Here, $\phi$ is the encoder and $\psi$ is the online predictor:
    \[
    \min_{\phi, \psi} \mathbb{E}_{(s_t, a_t, s_{t+1}) \sim \mathcal{D}} \left[ \| \psi(\phi(s_t), a_t) - \text{sg}(\phi(s_{t+1})) \|^2_2 \right].
    \]
    
    \item \textbf{Bootstrap Your Own Latent with Bidirectional Prediction (BYOL$\gamma$) \cite{lawson2025self}} captures long-range temporal consistency by predicting future states at horizons sampled geometrically. Here, $\psi_f$ is the forward predictor towards future states and $\psi_b$ is the backward predictor towards past states:
    \[
    \min_{\phi, \psi_f, \psi_b} \mathbb{E}_{\substack{(s_t, a) \sim \mathcal{D} \\ k \sim \text{Geom}(1-\gamma)}} \left[ \| \psi_f(\phi(s_t), a) - \text{sg}(\phi(s_{t+k})) \|^2_2 + \| \psi_b(\text{sg}(\phi(s_{t+k}))) - \phi(s_t) \|^2_2 \right].
    \]
\end{itemize}

\subsection{Forward-Backward Representations}
\label{appendix:bfms:fb}


Successor features require \(\phi\) to be designed before training and restrict expressivity to the linear span of those features. \cite{touati2021learning} address both shortcomings by replacing the feature-based decomposition with a direct low-rank factorization of the Successor Measure \cite{dayan1993improving, blier2021learning}. For a family of policies \(\{\pi_z\}_{z \in \mathcal{Z}}\) parameterized by task embeddings \(z \in \mathcal{Z} \subseteq \mathbb{R}^d\), the discounted successor measure of \(\pi_z\) starting from \((s, a)\) is
\begin{equation}
  M^{\pi_z}(s, a, X)
  \;=\; \sum_{t=0}^{\infty} \gamma^t\, \Pr\!\bigl(s_{t+1} \in X \mid s_0 = s,\, a_0 = a,\, \pi_z\bigr),
  \qquad X \subseteq \mathcal{S}.
  \label{eq:fb_succ_measure}
\end{equation}
The action-value function for any reward \(r\) can then be recovered via the linear functional \(Q^{\pi_z}_r(s,a) = \mathbb{E}_{s^+ \sim M^{\pi_z}(\cdot|s,a)}[r(s^+)]\), which makes the successor measure a sufficient statistic for all tasks simultaneously.

\paragraph{Training.}
Forward-Backward representations factorize the normalized successor measure into a bilinear product via two neural networks: a Forward map \(F : \mathcal{S} \times \mathcal{A} \times \mathcal{Z} \to \mathbb{R}^d\) and a Backward map \(B : \mathcal{S} \to \mathbb{R}^d\), trained so that \(F(s, a, z)^{\!\top} B(s') \approx M^{\pi_z}(s, a, \{s'\}) / \rho(s')\), where \(\rho\) is a reference state distribution.
The Bellman consistency of the successor measure, \(M^{\pi_z}(s,a, \cdot) = P(\cdot | s, a) + \gamma\, \mathbb{E}_{a' \sim \pi_z(\cdot|s')}[M^{\pi_z}(s', a', \cdot)]\), yields a tractable TD objective:
\begin{multline}
  \mathcal{L}_{\mathrm{FB}}(F, B)
  \;=\; \mathbb{E}_{\substack{(s,a,s') \sim \mathcal{D},\; z \sim \mathcal{Z} \\
                              a' \sim \pi_z(\cdot | s'),\; s^+ \sim \mathcal{D}}}
        \!\left( F(s, a, z)^{\!\top} B(s^+)
               - \gamma\, F(s', a', z)^{\!\top} B(s^+) \right)^2
  \\
  -\; 2\,\mathbb{E}_{(s,a,s') \sim \mathcal{D},\; z \sim \mathcal{Z}}
        \bigl[F(s, a, z)^{\!\top} B(s')\bigr],
  \label{eq:fb_loss}
\end{multline}
where the first term enforces Bellman consistency and the second anchors the diagonal of the successor measure (i.e., forces \(F(s,a,z)^{\!\top} B(s') = 1\) on transitions actually visited under \(\pi_z\)). At test time, a new reward \(r\) is encoded into a task embedding \(z_r = \mathbb{E}[r(s) B(s)]\) and the zero-shot policy is \(\pi_{z_r}(s) = \arg\max_a F(s, a, z_r)^{\!\top} z_r\), requiring no additional environment interaction.

\subsection{TD-JEPA}
\label{appendix:bfms:tdjepa}

FB learns the successor measure factorization directly in the state space, without any explicit state encoder. This makes it difficult to scale to raw pixel inputs, where compressed representations are essential, and prevents the reuse of learned features for representation-based transfer. Furthermore, FB offers no mechanism to learn a state encoder that is aligned with long-term, policy-conditioned dynamics. \cite{bagatella2025td} introduce TD-JEPA, which replaces the explicit successor measure factorization of FB with a latent-predictive objective inspired by the Joint-Embedding Predictive Architecture~(JEPA) paradigm of~\cite{lecun2022path}.

\paragraph{Training}
TD-JEPA trains four components end-to-end: 
a state encoder \(\phi : \mathcal{S} \to \mathbb{R}^{d_\phi}\), 
a task encoder \(\psi : \mathcal{S} \to \mathbb{R}^{d_\psi}\), 
a policy-conditioned predictor \(T_\phi : \mathbb{R}^{d_\phi} \times \mathcal{A} \times \mathcal{Z} \to \mathbb{R}^{d_\psi}\), 
and a family of parameterized policies \(\{\pi_z\}_{z \in \mathcal{Z}}\). 

The predictor is trained to approximate multi-step, policy-conditioned latent dynamics. TD-JEPA uses the Bellman equation for successor features in latent space, yielding the off-policy, single-step TD-JEPA loss:
\begin{equation}
  \mathcal{L}_{\mathrm{TD\text{-}JEPA}}(\phi, T_\phi)
  \;=\; \mathbb{E}_{\substack{(s,\,a,\,s') \sim \mathcal{D},\;\; z \sim \mathcal{Z} \\
                              a' \sim \pi_z(\cdot | s')}}
  \left\| T_\phi\!\bigl(\phi(s),\, a,\, z\bigr)
        - \bar\phi(s')
        - \gamma\; T_\phi\!\bigl(\bar\phi(s'),\, a',\, z\bigr)
  \right\|^2,
  \label{eq:tdjepa_loss}
\end{equation}
where \(\bar\phi\) denotes a stop-gradient copy of \(\phi\) (target encoder).

Zero-shot policy extraction follows the FB framework: at test time a, reward \(r\) is projected onto the task encoder as \(z_r = \mathbb{E}[r(s)\,\psi(s)]\), and the policy \(\pi_{z_r}(s) = \arg\max_a T_\phi(\phi(s), a, z_r)^{\!\top} z_r\) is executed without any environment interaction.

\subsection{BTD-FB}

Forward-Backward representations learn a rich, task-agnostic successor measure factorization, but their policy training component still relies on uniformly sampled task vectors $z \sim \text{Unif}(\mathcal{S}^{d-1})$. \cite{bendib2026improving} argue that this uniform sampling strategy suffers from a signal dilution effect: in high-dimensional latent spaces, uniformly drawn task vectors are nearly orthogonal to the behavioral space, causing the variance of returns across policies to vanish and the learning signal to collapse. BTD-FB addresses this by replacing the uniform task prior in FB with the Behavioral Task Distribution (BTD), a data-driven distribution fitted directly to the offline dataset.

\paragraph{Training.}
BTD-FB first trains Forward and Backward maps $F$ and $B$ using the standard FB objective (Eq.~\ref{eq:fb_loss}). Following training, the state embedding is fixed as $\phi(s) = \mathbb{E}[BB^\top]^{-1} B(s)$. To construct the BTD, $N_\tau$ sub-trajectories of random lengths are sampled from the offline dataset, and their empirical feature occupancies are computed and normalized to yield task vectors:
\begin{equation}
    z_\tau = \frac{\tilde{\psi}^\tau}{\|\tilde{\psi}^\tau\|_2}, \qquad \tilde{\psi}^\tau = \sum_{t=0}^{|\tau|} \gamma^t \phi(s_t).
\end{equation}
A Gaussian Mixture Model $p_\theta(z)$ is then fitted by maximum likelihood to the resulting empirical task set. The conditional policy is trained from scratch on these extracted tasks instead.

\newpage
\section{Background on Quality-Diversity Methods}
\label{app:qd_background}

Quality-Diversity~(QD) algorithms discover large collections of diverse, high-performing solutions simultaneously. A QD problem is defined by a fitness function \(r : \mathcal{S} \to \mathbb{R}\) and a behavioral descriptor \(\beta : \mathcal{S} \to \mathcal{B}\), where \(\mathcal{B}\) is discretized into a finite grid of cells. An archive \(\mathcal{C}\) stores at most one elite per cell, replaced only when a candidate achieves strictly higher fitness in the same cell. Performance is measured by \textit{coverage} (fraction of populated cells) and \textit{QD-score} (sum of all elite fitnesses). Most QD variants share this skeleton and 
differ only in their choice of variation operator, as summarized in 
Algorithm~\ref{alg:base_qd}.

\begin{algorithm}[H]
\caption{Base Quality-Diversity Algorithm}
\label{alg:base_qd}
\begin{algorithmic}[1]
\Require Fitness function $r$, descriptor $\phi$, budget $T$
\State Initialize archive $\mathcal{C} \leftarrow \emptyset$
\For{$t = 1$ \textbf{to} $T$}
    \State Sample parent $\theta \sim \mathcal{C}$ \Comment{randomly if $\mathcal{C} = \emptyset$}
    \State $\theta' \leftarrow \textsc{Operator}(\theta, \mathcal{C})$ \Comment{differs across methods}
    \State Evaluate fitness $f \leftarrow r(\theta')$ and descriptor $b \leftarrow \phi(\theta')$
    \If{$f > r(\mathcal{C}[b])$}
        \State $\mathcal{C}[b] \leftarrow (\theta',\, f)$
    \EndIf
\EndFor
\State \Return $\mathcal{C}$
\end{algorithmic}
\end{algorithm}

\subsection{MAP-Elites}

MAP-Elites (ME)~\cite{mouret2015illuminating} is the canonical QD algorithm. It addresses the exploration-exploitation tension in a single-objective optimizer by explicitly maintaining one best solution per behavioral cell, so that the search covers the descriptor space rather than collapsing onto a single high-fitness region.

\textbf{Operator.}
Parents are sampled from the archive and perturbed by an isotropic Gaussian mutation:
\begin{equation}
    \theta' = \theta + \varepsilon, \qquad \varepsilon \sim \mathcal{N}(0, \sigma^2 I).
    \label{eq:me_mutation}
\end{equation}
Each offspring is evaluated and inserted into its behavioral cell if it improves the current elite.

ME is simple and highly parallelizable. But because mutation is isotropic, ME does not exploit any structure of the fitness landscape: every direction in parameter space is equally likely to be explored, regardless of which directions have previously led to archive improvements.

\subsection{CMA-ME}

CMA-ME~\cite{fontaine2020covariance} addresses the sample-inefficiency of isotropic mutation by replacing it with CMA-ES~\cite{hansen2003reducing}, which maintains an adaptive covariance matrix \(\mathbf{C}\) estimated from directions that previously improved the archive.

\textbf{Operator.}
Offspring are sampled from a multivariate Gaussian whose covariance is updated from
archive improvement signals:
\begin{equation}
    \theta' = \theta + \varepsilon, \qquad \varepsilon \sim \mathcal{N}(0, \sigma^2 \mathbf{C}).
    \label{eq:cmame_mutation}
\end{equation}

Despite adapting the mutation distribution, CMA-ME still relies purely on undirected sampling: it does not use any task-specific gradient information to steer offspring towards higher-fitness regions.

\subsection{PGA-ME}

PGA-ME~\cite{nilsson2021policy} introduces a gradient-based variation operator to guide the search towards high-fitness regions, complementing random mutation with directed improvement steps.

\textbf{Operator.}
Half of each generation's offspring are produced by Gaussian mutation as in ME; the other half by gradient ascent on a learned critic \(Q_\phi\) trained via TD3:
\begin{equation}
    \theta \leftarrow \theta + \alpha \nabla_\theta Q_\phi\!\bigl(s,\, \pi_\theta(s)\bigr).
    \label{eq:pgame_grad}
\end{equation}
The critic is trained on transitions collected during archive evaluation.

The limitation with PGA-ME is that gradient steps optimize fitness without any awareness of the behavioral descriptor, so the method cannot explicitly target underpopulated regions of the archive, and exploration can collapse on the optimal behavior.

\subsection{DCRL-ME}

\textbf{Motivation.}
DCRL-ME~\cite{faldor2025synergizing} extends PGA-ME by conditioning both actor and critic on a target behavioral descriptor \(b^*\), so that gradient steps jointly optimize fitness \emph{and} steer the offspring toward a desired region of the descriptor space.

\textbf{Operator.}
The policy gradient update becomes descriptor-conditioned:
\begin{equation}
    \theta \leftarrow \theta + \alpha \nabla_\theta Q_\phi\!\bigl(s,\,
    \pi_\theta(s, b^*),\, b^*\bigr),
    \label{eq:dcrlme_grad}
\end{equation}
and the conditioned actor \(\pi_\theta(\cdot, b^*)\) is also used as a generative model, injecting candidates directly into target archive cells.

Both the actor and critic in DCRL-ME are trained from scratch using only transitions collected during archive evaluation, making them dependent on the quality and density of the reward signal encountered online.

\subsection{DDE-Elites}

\textbf{Motivation.}

DDE-Elites~\cite{gaier2020discovering} addresses the problem of high dimensionality of the policy parameter space by learning a compact Data-Driven Encoding (DDE) from the archive elites using a VAE, and using it as a variation operator within ME.

\textbf{Operators.}
At each generation, a VAE is retrained on all current archive solutions by minimzing this loss (maximizing the Evidence Lower Bound):
\begin{equation}
\mathcal{L}_\mathrm{VAE}
    = -\mathbb{E}_{q_{\theta_e}(z|\theta)}\!\left[\log p_{\theta_d}(\theta \mid z)\right]
    + \mathrm{KL}\!\left(q_{\theta_e}(z|\theta) \,\|\, \mathcal{N}(0,I)\right).
\end{equation}
Three variation operators are combined at each generation:
\textbf{(i) Isometric mutation} applies Gaussian noise directly in parameter space;
\textbf{(ii) line mutation} applies directional Gaussian noise, where the variance per dimension scales with the difference between two archive elites, biasing perturbations toward directions of known diversity;
\textbf{(iii) reconstructive crossover} shifts a solution toward the VAE-learned distribution without adding noise:
\begin{equation}
    \theta' = \tfrac{1}{2}\bigl(\theta + D_{\theta_d}(E_{\theta_e}(\theta))\bigr).
    \label{eq:dde_xover}
    \end{equation}
The mix of operators is selected at each generation by a UCB1 bandit algorithm~\cite{auer2002finite}, which tracks the success rate of each operator and balances exploration and exploitation automatically. 

The compact latent space simplifies search, but the VAE is trained only on solutions already in the archive, so the latent space does not necessarily reflect task-relevant structure.

\subsection{ME+Pretrain}
This is a baseline that we introduce to isolate the contribution of offline initialization from the structure of a behavioral latent space.
In ME+Pretrain, MLP policies are warm-started via offline TD3 on the same dataset as the BFM, using the task fitness as reward, before running standard ME.

\textbf{Operator.}
The pretraining objective is:
\begin{equation}
    \theta^{\star} = \arg\max_\theta\;
    \mathbb{E}_{(s,a) \sim \mathcal{D}}\!\left[
    Q_{\omega}\!\bigl(s,\, \pi_\theta(s)\bigr)\right]
    \label{eq:mepretrain}
\end{equation}
after which ME runs from \(\theta^{\star}\).

\newpage
\section{Experimental Details}
\label{app:experimental_details}

\subsection{Environments and Tasks}
\label{appendix:envs}

\begin{figure}[H]
    \centering
    \includegraphics[width=1.0\linewidth]{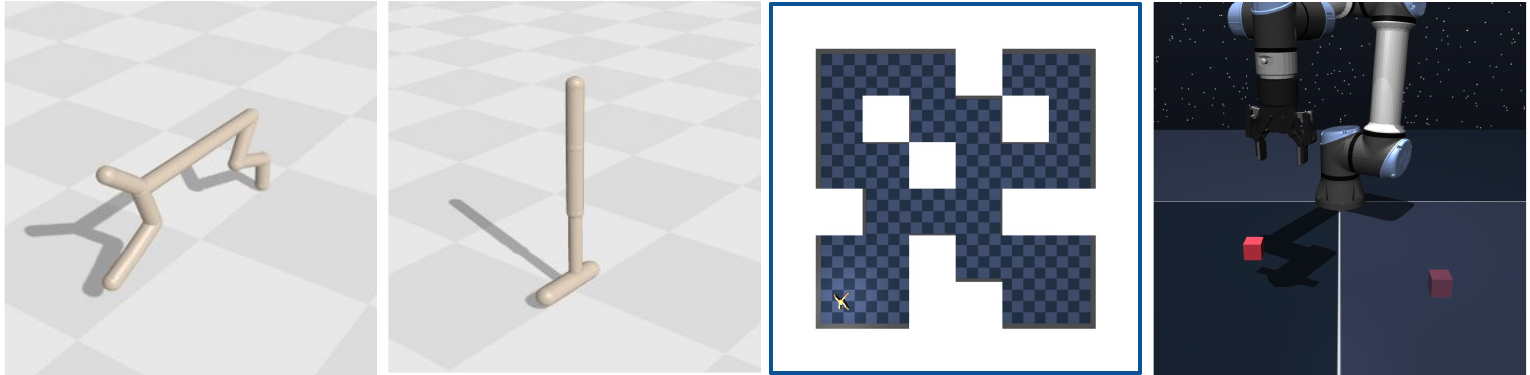}
    \caption{\textbf{Evaluation environments.} The four benchmarks used in this work: HalfCheetah\_uni, Walker2d\_uni, AntMaze-medium, and Cube-single (from left to right). Difficulty increases from left to right, with AntMaze-medium and Cube-single introducing long-horizon tasks and contact-rich interaction, respectively.}
    
    \label{fig:envs}
\end{figure}

\subsubsection*{HalfCheetah\_uni:} A planar 2D robot from QDax~\cite{chalumeau2024qdax}. It has a state dimension of 17 and an action dimension of 6. It represents the simplest setting with dense rewards and straightforward dynamics.

\textbf{Behavioral descriptor:}
\begin{itemize}
    \item \textbf{Foot contact pattern (2D):} The ratio of the contact of the front/back feet with the ground.
\end{itemize}
\textbf{Fitness functions:}
\begin{itemize}
    \item \textbf{Walk:} Reward is linearly proportional to forward velocity, capped at 2 m/s.
    \item \textbf{Run:} Reward is linearly proportional to forward velocity, capped at 5 m/s.
    \item \textbf{Walk backward:} Reward is linearly proportional to backward velocity, capped at -2 m/s.
    \item \textbf{Run backward:} Reward is linearly proportional to backward velocity, capped at -5 m/s.
    
\end{itemize}

\subsubsection*{Walker2D\_uni:} A bipedal walker from QDax, with slightly more complex dynamics than HalfCheetah\_uni. It has a state dimension of 17 and an action dimension of 6.

\textbf{Behavioral descriptor:}
\begin{itemize}
    \item \textbf{Foot contact pattern (2D):} The ratio of the contact of the front/back feet with the ground.
\end{itemize}
\textbf{Fitness functions:}
\begin{itemize}
    \item \textbf{Run forward}: Maximize the velocity of the center-of-mass of the Walker agent.
    \item \textbf{Run backward}: Maximize the backward velocity of the point-of-mass of the Walker agent.
    \item \textbf{Upright walk}: Maximize forward velocity while maintaining an upright torso above $0.3$.
    \item \textbf{Jump}: Maintain a center-of-mass height above $1.2$.
\end{itemize}

\subsubsection*{AntMaze-medium:} A quadrupedal Ant robot navigating a medium-sized maze from OGBench~\cite{park2024ogbench}. It has a state dimension of 29 and an action dimension of 8, and introduces long-horizon navigation with deceptive reward landscapes.

\textbf{Behavioral descriptor:}
\begin{itemize}
    \item \textbf{$(x,y)$ position:} the coordinate of the position of the ant in the maze.
    \item \textbf{Foot contact pattern (4D):} The ratio of the contact of the ant's feet with the ground.
\end{itemize}
\textbf{Fitness functions:}
\begin{itemize}
    \item \textbf{Energy efficiency}: Minimize arm energy consumption (norm of the actions), i.e. maximize the fitness function $f = \sqrt{d_A} - \|a_t\|_2$, with $d_A = 8$.
    \item \textbf{Reach center}: Reach the center of the maze in position (12, 12) with a radius of 1.
    \item \textbf{Top wall}: Reach the top wall of the maze after $x=20$.
    \item \textbf{Right wall}: Reach the right wall of the maze after $y=20$.
    \item \textbf{Top wall dense}: The dense version of \textbf{top wall}, with 
            $\text{fitness}= e^{(\text{ant.x} - 20) / 3} $
    \item \textbf{Right wall dense}: The dense version of \textbf{right wall}, with 
            $\text{fitness}= e^{(\text{ant.y} - 20) / 3} $

\end{itemize}

\paragraph{Cube-single:} A 6-DoF robotic arm from OGBench. The state space dimension is 28. The action space is 5-dimensional ($\Delta x, \Delta y, \Delta z, \Delta$yaw, gripper speed), and it represents the hardest setting requiring temporally extended contact-rich manipulation.

\textbf{Behavioral descriptor:}
\begin{itemize}
    \item \textbf{$(x,y)$ position}: the $(x,y)$ position of the cube.
    \item \textbf{$(x,z)$ position}: replaces the $y$ position with the height of the cube.
    \item \textbf{Grasp style}: (Relative grasp angle,  how wide the grip is).
    \item \textbf{Wrist configuration}: (the angle in joint\_3, the angle in  joint\_4)
\end{itemize}
\textbf{Fitness functions:}
\begin{itemize}
    \item \textbf{Energy efficiency}: Minimize arm energy consumption (norm of the actions), i.e. maximize the fitness function $f = \sqrt{d_A} - \|a_t\|_2$, with $d_A = 5$.
    \item \textbf{Pick-and-place}: Binary success of lifting and placing the cube at a target location.
\end{itemize}

\begin{table}[ht]
\centering
\caption{Summary of environments, behavioral descriptors, and fitness functions used in our
evaluation. Difficulty combines reward density and task complexity: \deasy\ = dense reward,
simple dynamics; \dmedium\ = dense reward, complex dynamics or navigation; \dhard\ = dense
but deceptive or long-horizon; \dvhard\ = sparse reward, contact-rich manipulation.}
\label{tab:envs}
\begin{tabular}{lllcc}
\toprule
\textbf{Environment} & \textbf{Behavioral Descriptor} & \textbf{Fitness Function} & \textbf{Density} & \textbf{Difficulty} \\
\midrule
\multirow{4}{*}{HalfCheetah}
    & \multirow{4}{*}{Foot contact} & Run backward  & Dense & \deasy \\
    &                               & Walk backward & Dense & \deasy \\
    &                               & Walk forward  & Dense & \deasy \\
    &                               & Run forward   & Dense & \deasy \\
\midrule
\multirow{4}{*}{Walker}
    & \multirow{4}{*}{Foot contact} & Run backward  & Dense & \deasy \\
    &                               & Run forward   & Dense & \deasy \\
    &                               & Upright walk  & Dense & \dmedium \\
    &                               & Jump          & Dense & \dmedium \\
\midrule
\multirow{8}{*}{AntMaze-medium}
    & $(x,y)$ position              & Energy efficiency & Dense  & \dmedium \\
    & \multirow{5}{*}{Foot contact} & Reach center      & Dense  & \dhard \\
    &                               & Reach right wall  & Dense  & \dhard \\
    &                               & Reach top wall    & Dense  & \dhard \\
    &                               & Reach right wall  & Sparse & \dvhard \\
    &                               & Reach top wall    & Sparse & \dvhard \\
\midrule
\multirow{5}{*}{Cube-single}
    & $(x,y)$ position of cube & Energy efficiency               & Dense  & \dhard \\
    & $(x,z)$ position of cube & Energy efficiency               & Dense  & \dhard \\
    & Grasp style              & Pick-and-place & Sparse & \dvhard \\
    & Wrist configuration      & Pick-and-place & Sparse & \dvhard \\
\bottomrule
\end{tabular}
\end{table}

\paragraph{QD-score normalization.}
Each environment contains multiple tasks with different reward scales, making raw QD-scores incomparable across tasks. To aggregate results within an environment, we apply per-task min-max normalization: for each task, we compute the minimum and maximum QD-score observed across all methods and all seeds, and rescale each method's score to $[0, 1]$ accordingly. Normalized scores are then averaged across all tasks within an environment to produce the aggregate curves shown in Figure~\ref{fig:main}. This normalization is applied identically to QD-score and max-fitness curves. Unnormalized per-task QD-scores are reported in Appendix~\ref{app:full_results}.

\subsection{Baselines Implementation Details}
\label{appendix:baseline_details}

All baselines share the same environment setup, archive configuration, training budget, and random seeds as BFM-QD for a fair comparison.

\begin{table}[H]
\centering
\caption{Shared hyperparameters across all methods.}
\label{tab:hp_shared}
\begin{tabular}{ll}
\toprule
\textbf{Parameter} & \textbf{Value} \\
\midrule
Archive grid size         & $50 \times 50$ \\
Number of generations     & 500 \\
Parallel environments     & 400 \\
Episode length (Brax)     & 500 steps \\
Episode length (OGBench)  & 1\,000 steps \\
\bottomrule
\end{tabular}
\end{table}

\paragraph{MAP-Elites.}
Offspring are produced by isotropic Gaussian mutation. Five operator instances of Gaussian mutation with a sigma ladder run in parallel, each handling an equal share of candidates per generation. Archive policies are 2-hidden-layer MLPs with ReLU activations: \texttt{input $\rightarrow$ 128 $\rightarrow$ relu $\rightarrow$ 128 $\rightarrow$ output $\rightarrow$ tanh}, taking the state observation as input and producing actions through a $\tanh$ output layer.

\begin{table}[H]
\centering
\caption{MAP-Elites hyperparameters.}
\label{tab:hp_me}
\begin{tabular}{ll}
\toprule
\textbf{Parameter} & \textbf{Value} \\
\midrule
Mutation $\sigma$  & $\{0.1,\, 0.5,\, 1.0,\, 1.0,\, 5.0\}$ \\
\bottomrule
\end{tabular}
\end{table}

\paragraph{CMA-ME.}
Uses the same emitter configuration and archive policy architecture as ME, but replaces isotropic mutation with CMA-ES, adapting the covariance matrix from archive improvement signals at each generation.

\paragraph{PGA-ME.}
Archive policies are 2-hidden-layer MLPs with 128 units per layer, ReLU activations, and a $\tanh$ output layer.
Half of each generation's offspring are produced by Gaussian mutation; the other half by one step of gradient ascent on a TD3 critic.
The critic is a twin Q-network with two hidden layers of 256 units and ReLU activations, taking the concatenation of state and action as input.
The TD3 actor has two hidden layers of 128 units with ReLU activations and produces a deterministic action through a $\tanh$ output layer.
Both are trained online from transitions collected during archive evaluation. 
A warmup phase is applied before any PG variation, allowing the replay buffer to fill before critic-guided mutations begin.
 
\begin{table}[H]
\centering
\caption{PGA-ME hyperparameters.}
\label{tab:hp_pgame}
\begin{tabular}{ll}
\toprule
\textbf{Parameter} & \textbf{Value} \\
\midrule
Archive policy MLP        & $2 \times 128$ hidden units \\
Critic hidden width       & 256 \\
Critic learning rate      & $3 \times 10^{-4}$ \\
Actor learning rate       & $3 \times 10^{-4}$ \\
Replay buffer size        & 1\,000\,000 \\
Batch size                & 256 \\
Discount $\gamma$         & 0.99 \\
Target smoothing $\tau$   & 0.005 \\
TD3 policy noise $\sigma$ & 0.2 \\
TD3 noise clip            & 0.5 \\
Target update frequency   & every 2 critic steps \\
Critic updates per gen.   & 256 \\
Warmup                    & 20 generations \\
\bottomrule
\end{tabular}
\end{table}

\paragraph{DCRL-ME.}
Extends PGA-ME with a descriptor-conditioned actor and critic, both taking the normalized 2-dimensional behavioral descriptor concatenated to their respective inputs (state for the actor, state-action for the critic).
The archive policy MLP, actor, and critic retain the same widths and activations as ME.
The offspring batch is split across three variation operators: 
Gaussian mutation (GA), descriptor-conditioned PG, and Actor Injection (AI).
For AI, a target descriptor $b^*$ is sampled uniformly from the descriptor space and the actor's first layer is analytically specialized to it: since the first layer computes $W[s;\, b^*] + b$, fixing $b^*$ reduces it to a standard affine map $W_s\, s + (W_{b^*} b^* + b)$, yielding a valid archive policy with no additional forward pass. TD3 training hyperparameters are otherwise identical to PGA-ME (Table~\ref{tab:hp_pgame}).

\begin{table}[H]
\centering
\caption{DCRL-ME hyperparameters. TD3 parameters are identical to PGA-ME (Table~\ref{tab:hp_pgame}).}
\label{tab:hp_dcrlme}
\begin{tabular}{ll}
\toprule
\textbf{Parameter} & \textbf{Value} \\
\midrule
Offspring split (GA / PG / AI) & 50\% / 25\% / 25\% \\
Similarity length scale $L$    & 0.1 \\
\bottomrule
\end{tabular}
\end{table}

\paragraph{DDE-Elites.}
Archive policies use the same MLP architecture as ME. The VAE has a symmetric encoder and decoder, each consisting of two hidden layers of 256 units with ReLU activations. The encoder maps a flattened policy parameter vector to a 50-dimensional latent space, producing a mean and log-variance from which a latent code is sampled via the reparameterization trick. The VAE is retrained every generation on all current archive weight vectors. Three variation operators are used: isometric mutation, line mutation, and reconstructive crossover (averaging a solution with its VAE reconstruction). Their mix is selected each generation by a UCB1 bandit. When fewer than the minimum number of elites are present, only isometric and line mutation are used.

\begin{table}[H]
\centering
\caption{DDE-Elites hyperparameters.}
\label{tab:hp_ddeelites}
\begin{tabular}{ll}
\toprule
\textbf{Parameter} & \textbf{Value} \\
\midrule
Archive policy MLP              & $2 \times 128$ hidden units \\
VAE latent dimension            & 50 \\
VAE encoder/decoder width       & 256 \\
VAE learning rate               & $3 \times 10^{-4}$ \\
VAE training epochs             & 5 \\
VAE KL weight                   & 0.01 \\
VAE retraining frequency        & every generation \\
Min.\ elites to retrain         & 50 \\
$\sigma$ isometric mutation     & 0.003 \\
$\sigma$ line mutation          & 0.1 \\
Bandit sliding window           & 1000 \\
\bottomrule
\end{tabular}
\end{table}

\paragraph{ME+Pretrain.}
MLP policies are initialized by offline TD3 training on the same dataset used to train the BFM, using the task fitness as reward. 
The offline actor and critic use the same architecture as in PGA-ME and DCRL-ME. The pretrained weights are then used as the starting point for ME with the same Gaussian mutation and emitter configuration as the ME baseline.

\begin{table}[H]
\centering
\caption{ME+Pretrain hyperparameters (offline TD3 phase).}
\label{tab:hp_mepretrain}
\begin{tabular}{ll}
\toprule
\textbf{Parameter} & \textbf{Value} \\
\midrule
Offline training steps        & 2\,000\,000 \\
Batch size                    & 2\,048 \\
Learning rate (actor, critic) & $10^{-4}$ \\
Target Polyak $\tau$          & 0.01 \\
Hidden width                  & 1024 \\
Feature dimension             & 512 \\
\bottomrule
\end{tabular}
\end{table}


\subsection{BFM Implementation Details}
\label{appendix:bfm_qd_details}

\paragraph{Laplacian (Lap).}
Lap learns only one state encoder $f$: $|\mathcal{S}| \xrightarrow{\texttt{ntanh}} 1024 \xrightarrow{\texttt{ReLU}} 1024 \xrightarrow{\texttt{ReLU}} d_z$.

\paragraph{BYOL$\gamma$ representation.}
BYOL$\gamma$ learns three networks: \\
$\bullet$ The state encoder $g$: $|\mathcal{S}| \xrightarrow{\texttt{ntanh}} 1024 \xrightarrow{\texttt{ReLU}} d_z$
\footnote{\texttt{ntanh} is layer normalization followed by a Tanh activation},
with a target network $\bar{g}$ updated by EMA. \\
$\bullet$ The forward predictor $f$: $(d_z+|\mathcal{A}|) \xrightarrow{\texttt{ntanh}} 1024 \xrightarrow{\texttt{ReLU}} 1024 \xrightarrow{\texttt{ReLU}} d_z$. \\
$\bullet$ The backward predictor $b$: $d_z \xrightarrow{\texttt{ntanh}} 1024 \xrightarrow{\texttt{ReLU}} 1024 \xrightarrow{\texttt{ReLU}} d_z$. \\
All three are trained jointly with an orthogonality regularizer on $g$.

\begin{table}[H]
\centering
\caption{SF-TD3 hyperparameters (shared across Lap and BYOL$\gamma$ variants).}
\label{tab:hp_sftd3}
\begin{tabular}{ll}
\toprule
\textbf{Parameter} & \textbf{Value} \\
\midrule
Latent dimension $d_z$           & 50 \\
SF training steps                & 2\,000\,000 \\
SF batch size                    & 1\,024 \\
SF learning rate                 & $10^{-4}$ \\
SF target Polyak $\tau$          & 0.01 \\
Actor/critic training steps      & 2\,000\,000 \\
Actor/critic learning rate       & $10^{-4}$ \\
TD3 batch size                   & 2\,048 \\
Discount $\gamma$                & 0.99 \\
Target EMA $\tau$             & 0.01 \\
z-inference batch size           & 10\,000 \\
\bottomrule
\end{tabular}
\end{table}

\paragraph{Forward-Backward (FB) model.}
The FB model follows \cite{touati2021learning}. The forward map $F$ takes $(s, a, z)$ through two parallel streams:
\\ $\bullet$ obs--action: $(|\mathcal{S}|+|\mathcal{A}|) \xrightarrow{\texttt{ntanh}} 512~[\texttt{ReLU}]$. \\
\\ $\bullet$ obs--$z$: $(|\mathcal{S}|+d_z) \xrightarrow{\texttt{ntanh}} 512~[\texttt{ReLU}]$. \\
Their concatenation passes through a shared trunk $1024 \xrightarrow{\texttt{ReLU}} 1024 \xrightarrow{\texttt{ReLU}} d_z$.

The backward network $B$ is a 4-hidden-layer MLP:
$|\mathcal{S}| \xrightarrow{\texttt{ntanh}} 1024 \xrightarrow{\texttt{ReLU}} 1024 \xrightarrow{\texttt{ReLU}} 1024 \xrightarrow{\texttt{ReLU}} d_z$,
$\ell_2$-normalized and rescaled to radius $\sqrt{d_z}$.
The actor shares the same preprocessed trunk as $F$ (obs-only and obs--$z$ streams) with a $\tanh$ action head.

\begin{table}[H]
\centering
\caption{FB hyperparameters.}
\label{tab:hp_bfmqd}
\begin{tabular}{ll}
\toprule
\textbf{Parameter} & \textbf{Value} \\
\midrule
Latent dimension $d_z$          & 50 \\
Training steps               & 2\,000\,000 \\
Batch size                   & 1\,024 \\
Learning rate                & $10^{-4}$ \\
Discount $\gamma$            & 0.99 \\
EMA $\tau$         & 0.01 \\
z-inference batch size              & 10\,000 \\
\bottomrule
\end{tabular}
\end{table}

\paragraph{TD-JEPA.}
TD-JEPA learns four components:
\\ $\bullet$ State encoder $\phi$: $|\mathcal{S}| \xrightarrow{\texttt{ntanh}} 1024 \xrightarrow{\texttt{ReLU}} d_\phi$, where $d_\phi = 512$.
\\ $\bullet$ Task encoder $\psi$:  $|\mathcal{S}| \xrightarrow{\texttt{ntanh}} 1024 \xrightarrow{\texttt{ReLU}} d_z$, $\ell_2$-normalized and rescaled to radius $\sqrt{d_z}$.
\\ $\bullet$ Forward predictor $T_\phi$ (predicting future $\psi$ from current $\phi$, twin MLP): $(d_\phi+d_z+|\mathcal{A}|) \xrightarrow{\texttt{ntanh}} 1024 \xrightarrow{\texttt{ReLU}} d_z$.
\\ $\bullet$ Backward predictor $T_\psi$ (predicting future $\phi$ from current $\psi$, twin MLP): $(d_z+d_z+|\mathcal{A}|) \xrightarrow{\texttt{ntanh}} 1024 \xrightarrow{\texttt{ReLU}} d_\phi$.

Each twin predictor averages its two heads before computing the TD-JEPA loss. Target networks are maintained for $\phi$, $\psi$, $T_\phi$, and $T_\psi$ via EMA.

$\bullet$ The actor: $(d_\phi+d_z) \xrightarrow{\texttt{ntanh}} 1024 \xrightarrow{\texttt{ReLU}} 1024 \xrightarrow{\texttt{ReLU}} |\mathcal{A}|$, with $\tanh$ output.

\begin{table}[H]
\centering
\caption{TD-JEPA hyperparameters.}
\label{tab:hp_tdjepa}
\begin{tabular}{ll}
\toprule
\textbf{Parameter} & \textbf{Value} \\
\midrule
Latent dimension $d_z$              & 50 \\
Training steps                      & 2\,000\,000 \\
Batch size                          & 1\,024 \\
Learning rate                       & $10^{-4}$ \\
Target EMA $\tau$                & 0.01 \\
Actor std.\ dev.\ $\sigma$          & 0.2 \\
Actor std.\ dev.\ clip              & 0.3 \\
Ortho.\ coeff.\ $\lambda_\phi$      & 1.0 \\
Ortho.\ coeff.\ $\lambda_\psi$      & 1.0 \\
z-inference batch size              & 10\,000 \\
\bottomrule
\end{tabular}
\end{table}

\paragraph{Backward Inference (BI) Operator.}
At each generation, half of the offspring are produced by Gaussian mutation and half by the BI operator. For the BI operator, state-reward pairs are sampled from the replay buffer and used to compute $z^*$ in closed form via Equation~\ref{eq:usf_inference}. Each parent is then interpolated toward $z^*$ with step size $\alpha$ and projected back onto the latent sphere.

\begin{table}[H]
\centering
\caption{Backward Inference (BI) operator hyperparameters.}
\label{tab:hp_bi}
\begin{tabular}{ll}
\toprule
\textbf{Parameter} & \textbf{Value} \\
\midrule
Step size $\alpha$              & 0.02 \\
BI probability                  & 0.5 \\
Gaussian mutation $\sigma$      & 1.0 \\
z-inference batch size          & 10\,000 \\
\bottomrule
\end{tabular}
\end{table}

\paragraph{Interpolation (Iso) Operator.}
At each generation, half of the offspring are produced by Gaussian mutation and half by the interpolation operator. For the interpolation operator, two parents are sampled from the archive and linearly interpolated with a coefficient $\alpha \sim \mathcal{U}[0,1]$, before projecting back onto the latent sphere.

\begin{table}[H]
\centering
\caption{Interpolation (Iso) operator hyperparameters.}
\label{tab:hp_iso}
\begin{tabular}{ll}
\toprule
\textbf{Parameter} & \textbf{Value} \\
\midrule
Interpolation probability       & 0.5 \\
$\alpha$ distribution           & $\mathcal{U}[0, 1]$ \\
Gaussian mutation $\sigma$      & 1.0 \\
\bottomrule
\end{tabular}
\end{table}

\paragraph{Dataset collection for BFM-QD}

For \texttt{HalfCheetah\_uni} and \texttt{Walker2D\_uni}, we collect our own dataset using RND \cite{burda2018exploration}. We use the official implementation from \cite{laskin2021urlb}. We keep the exact same hyperparameters.

\paragraph{Pseudo-algorithm of BFM-QD}
Algorithm~\ref{pseudo:FB-QD} summarizes the BFM-QD framework. The key difference from standard QD (Algorithm~\ref{alg:base_qd}) is that the archive stores latent codes $z$ rather than policy weights $\theta$, and all variation operators act directly in $\mathcal{Z}$. The BFM policy $\pi_{\text{BFM}}(\cdot \mid z)$ is called at evaluation time and remains frozen throughout.

\begin{algorithm}[H]
\caption{BFM-QD}
\label{pseudo:FB-QD}
\begin{algorithmic}[1]
\Require Pretrained BFM policy $\pi_{\text{BFM}}(\cdot \mid z)$
\Require Latent space $\mathcal{Z}$, archive $\mathcal{C}$
\Require Number of iterations $T$, batch size $N$
\State Initialize archive $\mathcal{C} \leftarrow \emptyset$
\For{$t = 1$ to $T$}
    \If{$\mathcal{C}$ is empty}
        \State Sample $N$ latent vectors $\{z_i\}_{i=1}^N \sim \mathcal{Z}$
    \Else
        \State Sample parent latent vectors $\{z_i\}_{i=1}^N$ from $\mathcal{C}$
    \EndIf

    \State Generate offspring latents: $z_i' = \text{Perturbation}(z_i)$

    \For{each latent vector $z_i'$}
        \State Roll out policy $\pi_{\text{BFM}}(\cdot \mid z_i')$
        \State Evaluate fitness $f_i$ and descriptor $b_i$
        \State Attempt insertion into archive $\mathcal{C}$ using $(z_i', f_i, b_i)$
    \EndFor
\EndFor
\State \Return archive $\mathcal{C}$
\end{algorithmic}
\end{algorithm}

Algorithms~\ref{pseudo:FB-BI} and~\ref{pseudo:FB-Iso} detail two variation operators:
\textbf{(1)} The backward Inference defined in \ref{method:bi};
\textbf{(2)} Interpolation defined in \ref{exp:interpolation}.

\begin{algorithm}[H]
\caption{BFM-QD with Backward Inference (BI) Operator}
\label{pseudo:FB-BI}
\begin{algorithmic}[1]
\Require Pretrained BFM policy $\pi_{\text{BFM}}(\cdot \mid z)$, feature map $\phi$
\Require Latent space $\mathcal{Z}$, archive $\mathcal{C}$, replay buffer $\mathcal{D}$
\Require Number of iterations $T$, batch size $N$, step size $\alpha$, latent dimension $d_z$
\State Initialize archive $\mathcal{C} \leftarrow \emptyset$, replay buffer $\mathcal{D} \leftarrow \emptyset$
\For{$t = 1$ to $T$}
    \If{$\mathcal{C}$ is empty}
        \State Sample $N$ latent vectors $\{z'_i\}_{i=1}^N \sim \mathcal{Z}$
    \Else
        \State Sample parent latent vectors $\{z_i\}_{i=1}^N$ from $\mathcal{C}$
        \State Sample $\mathcal{D}_r = \{(s_t, r(s_t))\}$ from $\mathcal{D}$
        \State $z^* \leftarrow \left(\mathbb{E}_{s \sim \mathcal{D}_r}[\phi(s)\phi(s)^\top] \right)^{-1} \mathbb{E}_{s \sim \mathcal{D}_r}[\phi(s) r(s)]$
        \For{each $z_i$}
            \If{with probability $0.5$}
                \State $z'_i \leftarrow z_i + \varepsilon,\quad 
                       \varepsilon \sim \mathcal{N}(0, \sigma^2 I)$ 
                       \Comment{Gaussian mutation}
            \Else
                \State $z'_i \leftarrow (1-\alpha) \cdot z_i + \alpha \cdot z^*$
                       \Comment{Backward Inference}
            \EndIf
            \State $z'_i \leftarrow z'_i / \|z'_i\| \cdot \sqrt{d_z}$
                   \Comment{project all offspring onto sphere}
        \EndFor
    \EndIf
    \State $z_i' \leftarrow z_i' / \|z_i'\| \cdot \sqrt{d_z}$ \Comment{project onto latent sphere}
    \For{each latent vector $z_i'$}
        \State Roll out policy $\pi_{\text{BFM}}(\cdot \mid z_i')$, collecting trajectory $\tau_i = \{(s_t, r(s_t))\}$
        \State Evaluate fitness $f_i$ and descriptor $b_i$
        \State Store trajectory $\tau_i$ in replay buffer $\mathcal{D}$
        \State Attempt insertion into archive $\mathcal{C}$ using $(z_i', f_i, b_i)$
    \EndFor
\EndFor
\State \Return archive $\mathcal{C}$
\end{algorithmic}
\end{algorithm}

\begin{algorithm}[H]
\caption{BFM-QD with Interpolation (Iso) Operator}
\label{pseudo:FB-Iso}
\begin{algorithmic}[1]
\Require Pretrained BFM policy $\pi_{\text{BFM}}(\cdot \mid z)$
\Require Latent space $\mathcal{Z}$, archive $\mathcal{C}$
\Require Number of iterations $T$, batch size $N$, latent dimension $d_z$
\State Initialize archive $\mathcal{C} \leftarrow \emptyset$
\For{$t = 1$ to $T$}
    \If{$\mathcal{C}$ is empty}
        \State Sample $N$ latent vectors $\{z'_i\}_{i=1}^N \sim \mathcal{Z}$
    \Else
        \State Sample parent latent vectors $\{z_i\}_{i=1}^N$ from $\mathcal{C}$
        \For{each $z_i$}
            \If{with probability $0.5$}
                \State $z'_i \leftarrow z_i + \varepsilon,\quad 
                       \varepsilon \sim \mathcal{N}(0, \sigma^2 I)$
                       \Comment{Gaussian mutation}
            \Else
                \State Sample second parent $z_j \sim \mathcal{C}$
                \State Sample $\alpha \sim \mathcal{U}[0,1]$
                \State $z'_i \leftarrow (1-\alpha) \cdot z_i + \alpha \cdot z_j$
                       \Comment{Interpolation}
            \EndIf
            \State $z'_i \leftarrow z'_i / \|z'_i\| \cdot \sqrt{d_z}$
                   \Comment{project all offspring onto sphere}
        \EndFor
    \EndIf
    \State $z_i' \leftarrow z_i' / \|z_i'\| \cdot \sqrt{d_z}$ \Comment{project onto latent sphere}
    \For{each latent vector $z_i'$}
        \State Roll out policy $\pi_{\text{BFM}}(\cdot \mid z_i')$
        \State Evaluate fitness $f_i$ and descriptor $b_i$
        \State Attempt insertion into archive $\mathcal{C}$ using $(z_i', f_i, b_i)$
    \EndFor
\EndFor
\State \Return archive $\mathcal{C}$
\end{algorithmic}
\end{algorithm}

\newpage
\section{Additional Results}
\label{app:additional_results}

\subsection{Expanded main results}

\begin{figure}[H]
    \centering
    \includegraphics[width=1\linewidth]{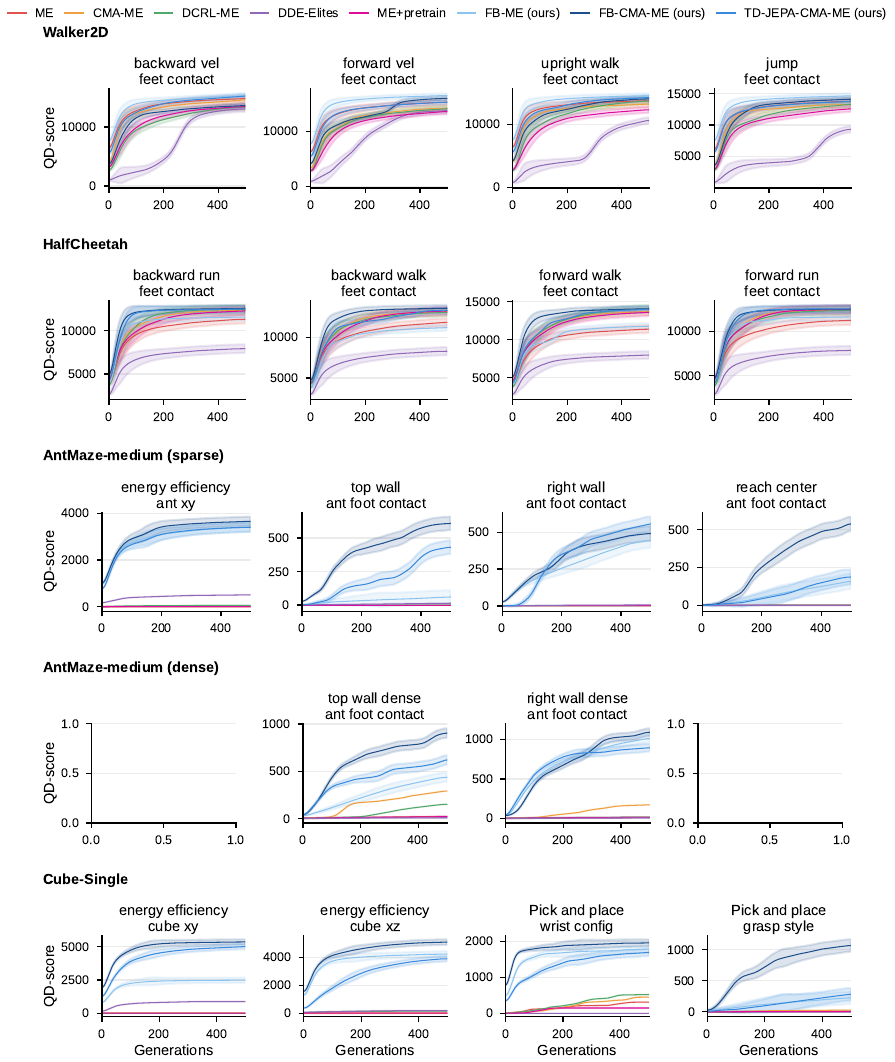}
    \caption{\textbf{QD-score across all environments and methods, averaged over 5 seeds.} BFM-QD variants (FB-ME, FB-CMA-ME, TD-JEPA-CMA-ME) consistently outperform parameter-space baselines (ME, CMA-ME, DCRL-ME, DDE-Elites), with margins that grow substantially in sparse and deceptive tasks (AntMaze-medium, Cube-single) where all parameter-space methods collapse to near-zero performance.}
    \label{fig:main_expanded}
\end{figure}

Figure~\ref{fig:main_expanded} reports the per-task, unnormalized QD-score curves for all methods across all 18 tasks and four environments. The results confirm the aggregated findings in Figure~\ref{fig:main}: BFM-QD variants consistently lead across tasks, with the performance gap over parameter-space baselines widening sharply as reward density decreases and task complexity increases.

\subsection{Archive Visualization}

\begin{figure}[H]
    \centering
    \includegraphics[width=1\linewidth]{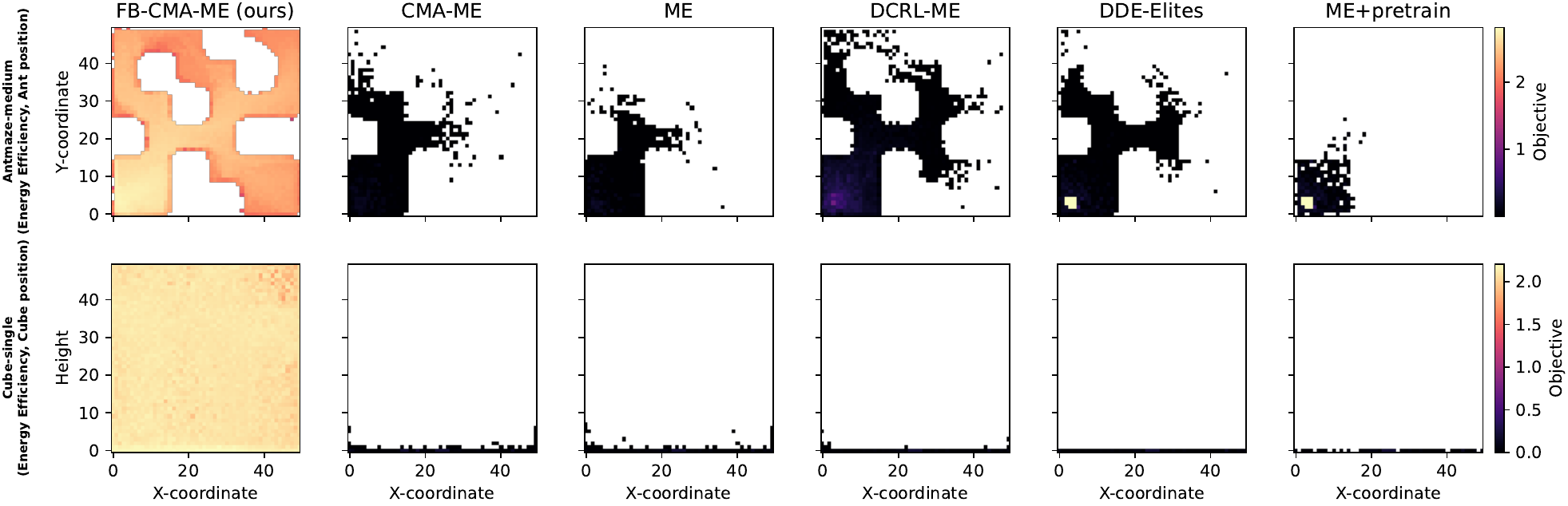}
    \caption{\textbf{Archive coverage.} Heatmaps of the descriptor space populated by FB-CMA-ME versus CMA-ME, ME, DCRL-ME, DDE-Elites, and ME+Pretrain, on AntMaze-medium (descriptor: $xy$-position, fitness: energy efficiency) and Cube-single (descriptor: cube $[x, \text{height}]$, fitness: energy efficiency). FB-CMA-ME achieves dense, broad coverage while parameter-space methods leave large regions entirely unpopulated.}
    \label{fig:heatmaps}
\end{figure}

Figure~\ref{fig:heatmaps} visualizes the archive heatmaps for FB-CMA-ME, CMA-ME, ME, DCRL-ME, and DDE-Elites on AntMaze-medium and Cube-single, making the collapse of parameter-space methods concrete. On AntMaze-medium, CMA-ME populates only a small cluster near the origin. On Cube-single, the CMA-ME archive is almost entirely empty, except for moving the cube on the surface of the table, because successfully holding the cube requires a temporally extended sequence of approach, grasp, and transport that random parameter perturbations cannot discover. FB-CMA-ME sidesteps both failure modes entirely, achieving dense and uniform coverage across almost the full descriptor space in both environments.

\subsection{The latent space is behaviorally rich by construction}
\label{appendix:random_sampling}

\begin{figure}[H]
    \centering
    \includegraphics[width=1\linewidth]{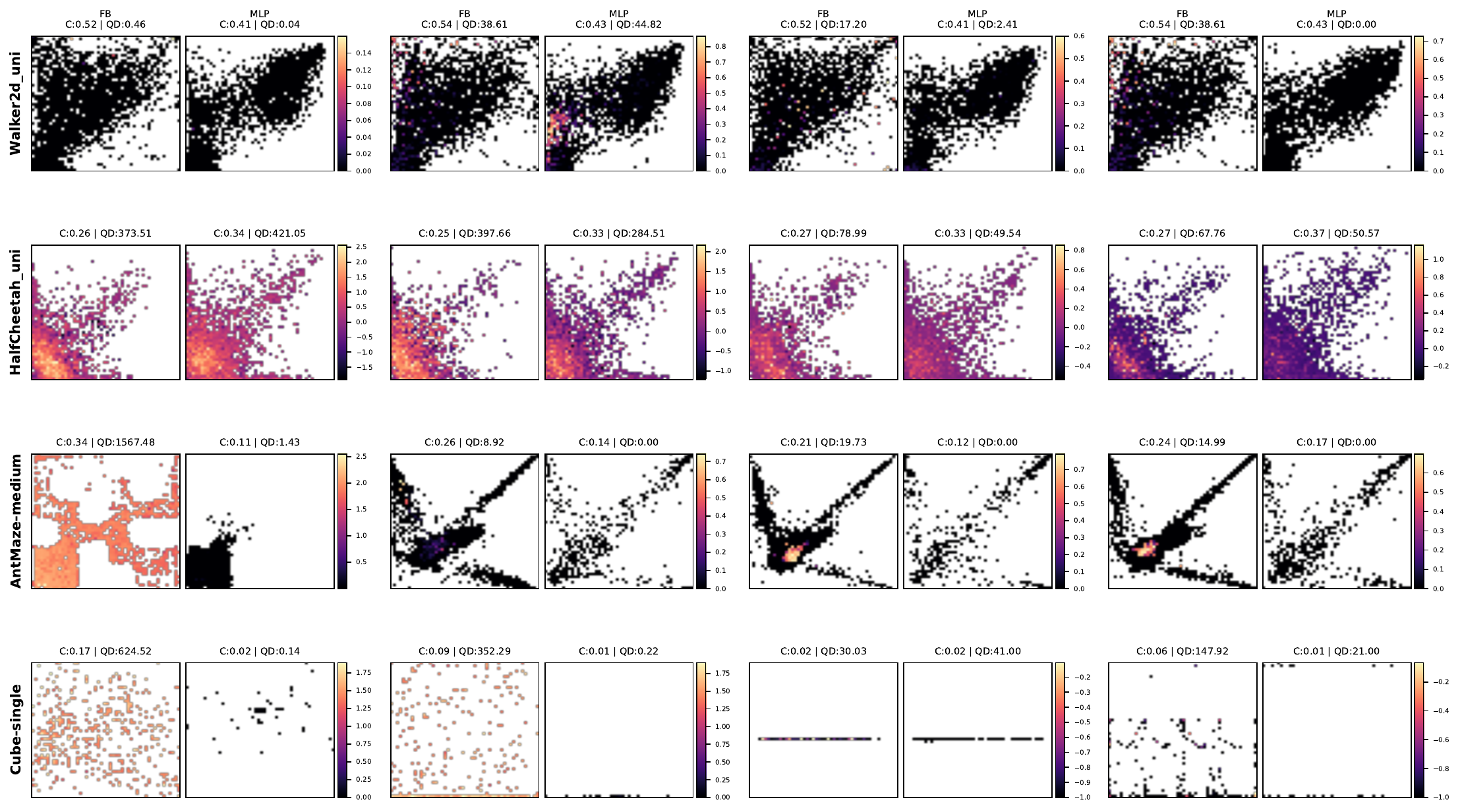}
    \caption{\textbf{Archive coverage from random sampling.}}
    \label{fig:random_sampling_archives}
    \vspace{-0.5cm}
\end{figure}

Figure~\ref{fig:random_sampling_archives} shows the archive coverage heatmaps obtained by random sampling from the trained FB latent space and the MLP parameter space, across all four environments.

\subsection{Effect of the Behavioral Foundation Model}
\label{app:bfm_comparison}

We investigate how the choice of BFM affects QD performance. Different BFMs learn the feature map $\phi$ and latent space $\mathcal{Z}$ through different objectives, which may yield latent spaces of varying quality and behavioral coverage. Understanding this sensitivity is important for practitioners choosing a BFM backbone for QD search.

\lead{Experiment} We fix the QD optimizer to CMA-ME and vary the BFM across five methods: Laplacian (Lap)~\cite{wu2018laplacian}, BYOL$\gamma$~\cite{lawson2025self}, TD-JEPA, BTD-FB~\cite{bendib2026improving}, and FB. For each BFM, we report its zero-shot performance on the task fitness function alongside the resulting QD-score to assess whether zero-shot performance is a reliable predictor of QD performance.

\begin{figure}[t]
    \centering
    \includegraphics[width=0.8\linewidth]{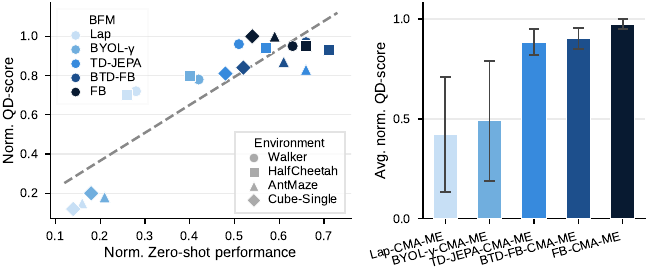}
    \caption{\textbf{BFM-QD does not require high zero-shot performance.} (Left) Methods with modest zero-shot returns still attain high QD-scores, suggesting latent space diversity matters more than zero-shot accuracy. (Right) FB-CMA-ME achieves the highest average QD-score, closely followed by BTD-FB-CMA-ME and TD-JEPA-CMA-ME. }
    \label{fig:bfm_comparison}
    \vspace{-0.5cm}
\end{figure}

\lead{Results} Results are shown in Figure~\ref{fig:bfm_comparison}. FB achieves the highest average QD-score, closely followed by BTD-FB and TD-JEPA, while Lap and BYOL$\gamma$ perform worse, particularly on AntMaze-medium and Cube-single. Overall, zero-shot performance is a good predictor of QD-score, and the regression confirms a clear positive correlation. However, on easy tasks, even BFMs with low zero-shot performance can achieve strong QD-scores, suggesting that a highly accurate BFM is not strictly necessary when tasks are dense and simple.

\begin{bluebox}
\lead{Takeaway:}
Zero-shot performance is a reliable proxy for QD performance: stronger BFMs consistently yield higher QD-scores. While weaker BFMs can still perform well on easy tasks, the choice of the BFM becomes critical as task difficulty increases.
\end{bluebox}

\subsection{Diversity as an Optimization Strategy}
\label{app:diversity}

We investigate whether diversity pressure in QD search provides benefits beyond single-objective optimization in the BFM latent space. A natural question is whether simply optimizing for fitness in $\mathcal{Z}$ is sufficient, or whether maintaining a diverse archive of solutions actively helps find better optima.

\lead{Experiment}
We compare three methods that all operate within the FB latent space. \textbf{FB Zero-shot} uses the closed-form backward inference (Equation~\ref{eq:usf_inference}) to obtain $z^*$ in a single pass without any additional environment interaction. \textbf{FB-CMA-ES} runs a single CMA-ES~\cite{hansen2016cma} optimizer directly in $\mathcal{Z}$, optimizing for maximum fitness without diversity pressure. \textbf{FB-CMA-ME} adds QD search on top of the same latent space.

\begin{figure}[H]
    \centering
    \includegraphics[width=0.8\linewidth]{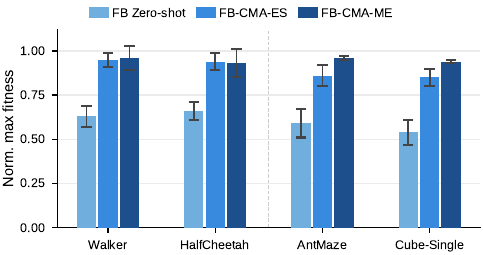}
    \caption{\textbf{Diversity as an optimization strategy.} Comparison of FB Zero-shot, FB-CMA-ES (single-objective optimization in $\mathcal{Z}$), and FB-CMA-ME (QD search in $\mathcal{Z}$). FB-CMA-ME outperforms FB-CMA-ES on maximum fitness, demonstrating that diversity pressure helps escape local optima during optimization.}
    \label{fig:ablation_prior}
    \vspace{-0.5cm}
\end{figure}

\lead{Results}
Results are shown in Figure~\ref{fig:ablation_prior}. FB-CMA-ME outperforms both FB-CMA-ES and FB Zero-shot consistently across all tasks. The gap between FB Zero-shot and FB-CMA-ES confirms that closed-form inference provides a useful initialization but not an optimal solution, due to approximation errors. More strikingly, FB-CMA-ME outperforms FB-CMA-ES even on maximum fitness: by maintaining solutions across diverse behavioral niches, diversity pressure helps escape local optima that a focused single-objective optimizer converges to.

\begin{bluebox}
\lead{Takeaway:}
Diversity pressure is not only beneficial for coverage: it also improves maximum fitness by preventing premature convergence to local optima in the latent space.
\end{bluebox}

\subsection{Sensitivity to the Offline Dataset Quality}
\label{app:dataset}

We investigate how the quality and composition of the offline dataset used to pretrain the BFM affect QD performance. Since the diversity of the latent space is ultimately bounded by the diversity of the training data, dataset composition is a critical factor for BFM-QD practitioners.

\lead{Experiment}
We vary the dataset composition on AntMaze-medium across six conditions: a \textbf{Biased} dataset restricted to the lower-left region of the maze, and five mixtures interpolating between fully random and expert data (\textbf{Random}, \textbf{75R/25E}, \textbf{50R/50E}, \textbf{25R/75E}, \textbf{Expert}). All other hyperparameters are fixed, and the QD-score is averaged across all AntMaze-medium tasks.

\begin{figure}[H]
    \centering
    \includegraphics[width=0.9\linewidth]{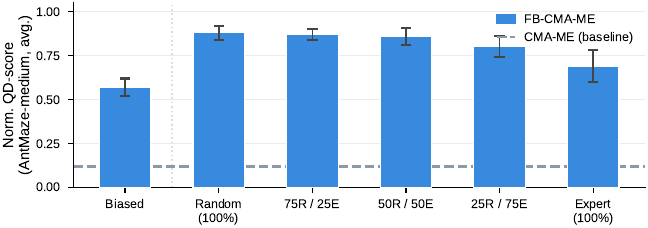}
    \caption{\textbf{Sensitivity to offline dataset quality.} Normalized QD-score on AntMaze-medium for FB-CMA-ME trained on datasets of varying composition: a spatially biased dataset (lower-left region only), and five mixtures of random and expert data. Performance degrades as expert data replaces random data, confirming that behavioral coverage matters more than trajectory quality for BFM-QD. Even the biased dataset substantially outperforms the CMA-ME baseline (dashed line).}
    \label{fig:dataset_sensitivity}
    \vspace{-0.5cm}
\end{figure}

\lead{Results}
Results are shown in Figure~\ref{fig:dataset_sensitivity}. Performance degrades monotonically as expert data replaces random data: expert trajectories cover only a narrow set of optimal paths, reducing the behavioral diversity of the latent space. This confirms that coverage matters more than trajectory quality for BFM-QD. Notably, even the spatially biased dataset does not collapse and substantially outperforms CMA-ME, suggesting that the BFM learns environment dynamics and motor primitives that generalize beyond the covered region.

\begin{bluebox}
\lead{Takeaway:}
Behavioral coverage in the offline dataset matters more than trajectory quality: random exploration datasets provide a richer and more diverse behavioral prior than expert ones. Even a spatially restricted random-exploration dataset yields a functional latent space that supports strong QD performance.
\end{bluebox}

\subsection{Additional Latent-Space Search Baselines}
\label{app:pretrain_latent}

DDE-Elites searches a learned latent space but learns it online from the archive, and ME+Pretrain uses offline data but searches parameter space. We test baselines that combine both, and Policy Manifold Search (PoMS)~\cite{rakicevic2021policy}, which runs ME in the latent space of an autoencoder over policy parameters.

\textbf{Experiment} We train three baselines:
\begin{itemize}
    \item \textbf{DDE-Elites+Pretrain (TD3)} pretrains 10 independent TD3 policies offline on the BFM dataset, using the task fitness as reward (as in ME+Pretrain). A single converged TD3 run collapses to one behavior and gives the VAE nothing to encode diversity from, so we keep 10 intermediate checkpoints per run (100 policies). This population initializes the archive, and we then run DDE-Elites.
    \item \textbf{DDE-Elites+Pretrain (BC)} follows the same protocol, but policies are pretrained by behavioral cloning on dataset trajectories, since different trajectories represent different behaviors.
    \item \textbf{PoMS} is closely related to DDE-Elites, as both encode archive solutions with an autoencoder over policy parameters.
\end{itemize}
We evaluate on Walker and Cube-single over 5 seeds. 

\textbf{Results} Table~\ref{tab:pretrain_latent} reports normalized QD-score. FB-ME outperforms all baselines on both environments. On Walker, pretraining does not help: DDE-Elites+Pretrain reaches $0.61$ (BC) and $0.76$ (TD3), versus $0.74$ for DDE-Elites, and none of the DDE-Elites or PoMS variants reaches ME. On Cube-single, pretraining raises DDE-Elites from $0.09$ to $0.15$ (BC) and $0.14$ (TD3), which remains far below FB-ME ($0.58$). PoMS falls below DDE-Elites on both environments ($0.63$ and $0.07$). These baselines imitate trajectories, optimize a single reward, or learn the latent space from the policy parameters in the archive, so their policies are bounded by the dataset, by one objective, or by the archive itself. A BFM instead learns the environment dynamics and trains policies to explore the resulting feature space.

\begin{table}[h]
\centering\small
\caption{Normalized QD-score (mean $\pm$ std, 5 seeds).}
\label{tab:pretrain_latent}
\begin{tabular}{lcc}
\toprule
Method & Walker & Cube \\
\midrule
ME & $0.93\pm0.02$ & $0.02\pm0.01$ \\
DDE-Elites & $0.74\pm0.21$ & $0.09\pm0.03$ \\
ME+Pretrain & $0.81\pm0.11$ & $0.02\pm0.00$ \\
DDE-Elites+Pretrain (BC) & $0.61\pm0.09$ & $0.15\pm0.04$ \\
DDE-Elites+Pretrain (TD3) & $0.76\pm0.13$ & $0.14\pm0.03$ \\
PoMS & $0.63\pm0.13$ & $0.07\pm0.03$ \\
FB-ME (ours) & $\mathbf{0.98\pm0.01}$ & $\mathbf{0.58\pm0.19}$ \\
\bottomrule
\end{tabular}
\end{table}

\begin{bluebox}
\textbf{Takeaway:} Neither offline pretraining nor a latent space learned over policy parameters is sufficient. The gains come from the structure of the BFM latent space.
\end{bluebox}

\subsection{Backward Inference vs.\ Sampling Around the Inferred Optimum}
\label{app:around}

BI moves each parent toward $z^*$. A simpler alternative is to sample offspring around $z^*$.

\textbf{Experiment} We sample $z'=z^*+\epsilon$ with $\epsilon\sim\mathcal{N}(0,\sigma^2 I)$ for $\sigma\in\{0.5,1.0\}$, with $z^*$ computed from the pretraining dataset to isolate the effect of data quality early in the search. We compare to FB-BI on Walker and Cube-single over 3 seeds.

\textbf{Results} Table~\ref{tab:around} reports normalized QD-score. Sampling around a fixed point confines the search to a small region of the latent space with behaviors close to that of $z^*$, instead of exploring the whole behavioral space. FB-BI performs far better because each parent keeps its own behavior and moves only a small step toward $z^*$.

\begin{table}[h]
\centering\small
\caption{Normalized QD-score (mean $\pm$ std, 3 seeds).}
\label{tab:around}
\begin{tabular}{lcc}
\toprule
Method & Walker & Cube \\
\midrule
$z^*+\epsilon$, $\sigma=0.5$ & $0.16\pm0.03$ & $0.09\pm0.01$ \\
$z^*+\epsilon$, $\sigma=1.0$ & $0.18\pm0.02$ & $0.09\pm0.02$ \\
FB-BI & $\mathbf{0.92\pm0.04}$ & $\mathbf{0.96\pm0.05}$ \\
\bottomrule
\end{tabular}
\end{table}

\begin{bluebox}
\textbf{Takeaway:} BI works because it moves each solution individually toward $z^*$ while preserving diversity. Sampling around $z^*$ does not.
\end{bluebox}

\subsection{Pretraining Data Collected by QD}
\label{app:qddata}

We use RND to collect pretraining data. QD could also collect it, but this commits the dataset to a descriptor, whereas RND is descriptor-agnostic.

\textbf{Experiment} We retrain FB on data collected by our own FB-QD search, and compare the resulting QD-score to FB trained on RND data. We evaluate on Walker (forward velocity) and AntMaze (top wall).

\textbf{Results} Table~\ref{tab:qddata} shows that QD-collected data reduces performance. Once QD finds a high-fitness elite for a cell, it refines solutions near that elite instead of covering new transitions. RND instead visits states with high prediction error and covers the state space. BFM pretraining needs data that covers the transition dynamics broadly, not only along the descriptor axes.

\begin{table}[h]
\centering\small
\caption{QD-score of FB-ME by pretraining data source (mean $\pm$ std).}
\label{tab:qddata}
\begin{tabular}{lcc}
\toprule
Pretraining data & Walker & AntMaze \\
\midrule
RND & $\mathbf{16471\pm175}$ & $\mathbf{61\pm45}$ \\
QD (FB-QD) & $5654\pm12$ & $46\pm9$ \\
\bottomrule
\end{tabular}
\end{table}

\begin{bluebox}
\textbf{Takeaway:} Broad coverage of the dynamics matters more than diversity along the test descriptor. Exploration data beats goal-directed data.
\end{bluebox}

\subsection{Sensitivity to the Backward Inference Step Size}
\label{app:alpha}

\textbf{Experiment} We run FB-BI on Walker with step size $\alpha\in\{0, 0.005, 0.02, 0.05, 0.1, 0.2, 0.5, 1\}$ over 3 seeds. $\alpha=0$ recovers FB-ME, and $\alpha=1$ replaces each solution with $z^*$. Our default is $\alpha=0.02$.

\begin{figure}[ht]
    \centering
    \includegraphics[width=0.5\linewidth]{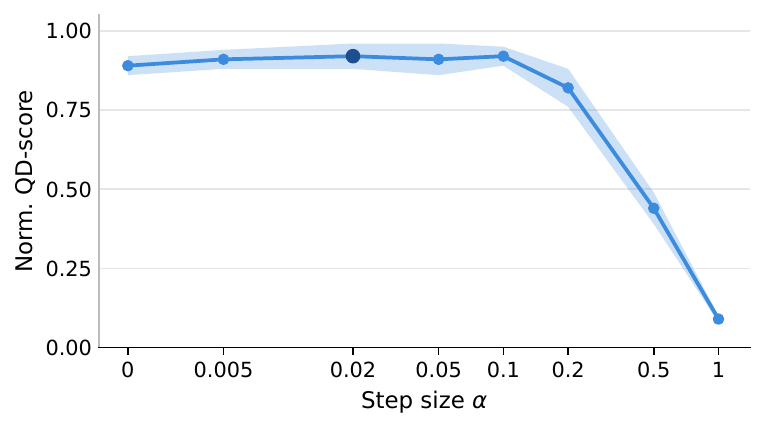}
    \caption{Normalized QD-score of FB-BI on Walker vs.\ step size $\alpha$ (mean $\pm$ std, 3 seeds).}
    \label{fig:alpha_ablation}
\end{figure}

\textbf{Results} Figure~\ref{fig:alpha_ablation} reports normalized QD-score. Performance is stable for $\alpha\in[0.005,0.1]$, and our default lies inside this plateau. Beyond $\alpha=0.1$ it degrades, since larger $\alpha$ pulls offspring toward the single point $z^*$ and collapses diversity (Section~4.2).

\begin{bluebox}
\textbf{Takeaway:} FB-BI is robust to $\alpha$ over a wide range. Large steps collapse diversity.
\end{bluebox}

\subsection{Compute Resources}
\label{app:compute}

All experiments were run on NVIDIA H100 GPUs. Table~\ref{tab:compute} reports wall-clock times per run. BFM pretraining is a one-time cost shared across all tasks within a given environment, so the amortized cost per QD run is dominated by the search itself.

\begin{table}[H]
\centering
\caption{Compute cost per run on a single NVIDIA H100 GPU.
  BFM pretraining is performed once per environment and reused across
  all tasks and seeds.}
\label{tab:compute}
\begin{tabular}{llr}
\toprule
\textbf{Phase} & \textbf{Method} & \textbf{GPU-hours} \\
\midrule
\multirow{5}{*}{BFM Pretraining (one-time)}
  & BYOL$\gamma$ & 2.67 \\
  & Lap     & 2.67 \\
  & TD-JEPA       & 2.50 \\
  & FB            & 2.32 \\
  & BTD-FB        & 3.17 \\
\midrule
\multirow{8}{*}{QD Search (per run)}
  & FB-ME          & 1.63 \\
  & FB-CMA-ME      & 1.83 \\
  & TD-JEPA-CMA-ME & 1.87 \\
  & ME             & 1.28 \\
  & CMA-ME         & 1.47 \\
  & DCRL-ME        & 2.33 \\
  & DDE-Elites         & 2.25 \\
  & ME+Pretrain    & 1.28 \\
\bottomrule
\end{tabular}
\end{table}


\newpage

\newpage
\subsection{Results Table}
\label{app:full_results}

Tables \ref{tab:results_walker2d}–\ref{tab:results_cube} report the final unnormalized QD-scores (mean ± std over 5 seeds) for all methods across Walker2D, HalfCheetah, AntMaze-medium, and Cube-single.

\begin{table}[ht]
\centering
\small
\setlength{\tabcolsep}{7pt}
\caption{QD-scores on \textbf{Walker2D} (mean\,$\pm$\,std over 5 seeds).
  \textbf{Bold}: best; \underline{underline}: second best.}
\label{tab:results_walker2d}
\begin{tabular}{lcccc}
\toprule
\textbf{Method} & \textbf{Backward Vel} & \textbf{Forward Vel} & \textbf{Upright Walk} & \textbf{Jump} \\
\midrule
\multicolumn{5}{l}{\textit{Baselines}} \\[1pt]
\rowcolor{baselinebg}
MAP-Elites      & 14801$\pm$514 & 15253$\pm$538 & 13746$\pm$493 & 13772$\pm$465 \\
\rowcolor{baselinebg}
CMA-ME          & 14580$\pm$489 & 14177$\pm$507 & 13241$\pm$511 & 13262$\pm$481 \\
\rowcolor{baselinebg}
PGA-ME          & 12410$\pm$491 & 13081$\pm$535 & 12124$\pm$474 & 12456$\pm$475 \\
\rowcolor{baselinebg}
DCRL-ME         & 13410$\pm$413 & 14089$\pm$456 & 13714$\pm$436 & 13286$\pm$462 \\
\rowcolor{baselinebg}
ME+Pretrain     & 13465$\pm$268 & 13576$\pm$264 & 12328$\pm$280 & 12595$\pm$268 \\
\rowcolor{baselinebg}
DDE-Elites          &  13097$\pm$105 &  13723$\pm$132   &  10742$\pm$210   &  9388$\pm$165   \\
\addlinespace[3pt]
\multicolumn{5}{l}{\textit{BFM-QD (ours)}} \\[1pt]
\rowcolor{ours!15}
FB-ME           & 15307$\pm$219 & 16471$\pm$175 & 14533$\pm$198 & 14641$\pm$228 \\
\rowcolor{ours!15}
FB-CMA-ME       & 13673$\pm$169 & 15986$\pm$169 & 14051$\pm$230 & 14115$\pm$188 \\
\rowcolor{ours!15}
TD-JEPA-CMA-ME  & 15302$\pm$272 & 15291$\pm$277 & 14268$\pm$225 & 13747$\pm$225 \\
\rowcolor{ours!15}
Lap-CMA-ME          &  9358$\pm$166 & 10391$\pm$154 &  9994$\pm$227 & 10043$\pm$192 \\
\rowcolor{ours!15}
BYOL$\gamma$-CMA-ME         &  9401$\pm$176 &  9932$\pm$187 & 10225$\pm$220 &  9473$\pm$204 \\
\rowcolor{ours!15}
BTD-FB-CMA-ME   & 16135$\pm$220 & 17546$\pm$183 & 13350$\pm$200 & 13307$\pm$221 \\
\rowcolor{ours!15}
FB-CMA-ME-Iso      & 13432$\pm$174 & 17087$\pm$175 & 15365$\pm$220 & 14291$\pm$194 \\
\rowcolor{ours!15}
FB-BI    & \best{17807$\pm$227} & \best{18908$\pm$180} & \second{15474$\pm$188} & \second{15328$\pm$225} \\
\rowcolor{ours!15}
FB-PGA-ME       & \second{16212$\pm$218} & \second{17906$\pm$186} & \best{16368$\pm$187} & \best{16714$\pm$225} \\
\bottomrule
\end{tabular}
\end{table}

\begin{table}[ht]
\centering
\small
\setlength{\tabcolsep}{7pt}
\caption{QD-scores on \textbf{HalfCheetah} (mean\,$\pm$\,std over 5 seeds).
  \textbf{Bold}: best; \underline{underline}: second best.}
\label{tab:results_halfcheetah}
\begin{tabular}{lcccc}
\toprule
\textbf{Method} & \textbf{Backward Run} & \textbf{Backward Walk} & \textbf{Forward Walk} & \textbf{Forward Run} \\
\midrule
\multicolumn{5}{l}{\textit{Baselines}} \\[1pt]
\rowcolor{baselinebg}
MAP-Elites      & 11331$\pm$524 & 11874$\pm$534 & 11402$\pm$490 & 11230$\pm$520 \\
\rowcolor{baselinebg}
CMA-ME          & 12381$\pm$507 & 13200$\pm$514 & 13724$\pm$469 & 12371$\pm$534 \\
\rowcolor{baselinebg}
PGA-ME          & 10685$\pm$463 & \second{14326$\pm$473} & 12523$\pm$478 & 12636$\pm$498 \\
\rowcolor{baselinebg}
DCRL-ME         & 12466$\pm$460 & 13103$\pm$479 & 14082$\pm$456 & 12324$\pm$472 \\
\rowcolor{baselinebg}
ME+Pretrain     & 12302$\pm$291 & 13293$\pm$338 & 13618$\pm$315 & 12425$\pm$309 \\
\rowcolor{baselinebg}
DDE-Elites          &  7931$\pm$120   &  8311$\pm$125   &  7981$\pm$216   &  7860$\pm$198   \\
\addlinespace[3pt]
\multicolumn{5}{l}{\textit{BFM-QD (ours)}} \\[1pt]
\rowcolor{ours!15}
FB-ME           & 11589$\pm$228 & 11246$\pm$228 & 11781$\pm$224 & 11590$\pm$239 \\
\rowcolor{ours!15}
FB-CMA-ME       & 12574$\pm$165 & 13617$\pm$201 & \second{14096$\pm$203} & 12470$\pm$223 \\
\rowcolor{ours!15}
TD-JEPA-CMA-ME  & 12502$\pm$234 & 13387$\pm$229 & 14002$\pm$220 & 12495$\pm$282 \\
\rowcolor{ours!15}
Lap-CMA-ME          &  7341$\pm$156 &  8051$\pm$189 &  9281$\pm$184 &  7952$\pm$232 \\
\rowcolor{ours!15}
BYOL$\gamma$-CMA-ME         &  9055$\pm$159 &  9849$\pm$215 & 11214$\pm$195 &  8557$\pm$239 \\
\rowcolor{ours!15}
BTD-FB-CMA-ME   & 10578$\pm$220 & 10488$\pm$220 & 10573$\pm$217 & 11394$\pm$230 \\
\rowcolor{ours!15}
FB-CMA-ME-Iso      & \best{14315$\pm$165} & \best{14839$\pm$200} & \best{14745$\pm$212} & \second{13232$\pm$230} \\
\rowcolor{ours!15}
FB-BI    & 12890$\pm$233 & 12514$\pm$221 & 12360$\pm$224 & 13077$\pm$247 \\
\rowcolor{ours!15}
FB-PGA-ME       & \second{14105$\pm$233} & 12520$\pm$219 & 12960$\pm$223 & \best{13745$\pm$240} \\
\bottomrule
\end{tabular}
\end{table}

\newpage
\begin{table}[ht]
\centering
\small
\caption{QD-scores on \textbf{AntMaze-Medium} (mean\,$\pm$\,std over 5 seeds).
  \textbf{Bold}: best; \underline{underline}: second best.
  Superscript\textsuperscript{D} denotes the dense-reward variant.}
\label{tab:results_antmaze_medium}
\adjustbox{max width=\textwidth}{%
\begin{tabular}{lcccccc}
\toprule
\textbf{Method}
  & \textbf{Energy Eff.}
  & \textbf{Top Wall}
  & \textbf{Right Wall}
  & \textbf{Reach Center}
  & \textbf{Top Wall\textsuperscript{D}}
  & \textbf{Right Wall\textsuperscript{D}} \\
\midrule
\multicolumn{7}{l}{\textit{Baselines}} \\[1pt]
\rowcolor{baselinebg}
MAP-Elites      &    6$\pm$0  &   0$\pm$0  &   0$\pm$0  &   0$\pm$0  &   23$\pm$0  &   14$\pm$0 \\
\rowcolor{baselinebg}
CMA-ME          &    6$\pm$0  &   0$\pm$0  &   0$\pm$0  &   0$\pm$0  &  293$\pm$0  &  175$\pm$0 \\
\rowcolor{baselinebg}
PGA-ME          &   22$\pm$2  &   0$\pm$0  &   0$\pm$0  &   0$\pm$0  &   76$\pm$0  &    5$\pm$1 \\
\rowcolor{baselinebg}
DCRL-ME         &   61$\pm$0  &   0$\pm$0  &   0$\pm$0  &   0$\pm$0  &  153$\pm$0  &   17$\pm$0 \\
\rowcolor{baselinebg}
ME+Pretrain     &    5$\pm$0  &   0$\pm$0  &   0$\pm$0  &   0$\pm$0  &   20$\pm$0  &   13$\pm$0 \\
\rowcolor{baselinebg}
DDE-Elites          &  509$\pm$0  &  14$\pm$3  &   7$\pm$4  &   0$\pm$0  &    8$\pm$1  &    7$\pm$4 \\
\addlinespace[3pt]
\multicolumn{7}{l}{\textit{BFM-QD (ours)}} \\[1pt]
\rowcolor{ours!15}
FB-ME           & 3406$\pm$52  &   61$\pm$45  &  446$\pm$51  &  161$\pm$51  &  441$\pm$52  & 1018$\pm$57 \\
\rowcolor{ours!15}
FB-CMA-ME       & 3657$\pm$51  & \second{611$\pm$50}  &  498$\pm$49  & \second{552$\pm$49}  & \second{909$\pm$49}  & 1099$\pm$49 \\
\rowcolor{ours!15}
TD-JEPA-CMA-ME  & 3407$\pm$88  &  436$\pm$10  & \second{565$\pm$56}  &  192$\pm$21  &  638$\pm$66  &  900$\pm$50 \\
\rowcolor{ours!15}
Lap-CMA-ME          &  274$\pm$60  &   14$\pm$37  &  105$\pm$53  &   14$\pm$42  &   94$\pm$46  &  133$\pm$66 \\
\rowcolor{ours!15}
BYOL$\gamma$-CMA-ME         &  947$\pm$54  &   10$\pm$51  &   59$\pm$51  &   36$\pm$50  &   64$\pm$52  &  151$\pm$58 \\
\rowcolor{ours!15}
BTD-FB-CMA-ME   & 3637$\pm$55  &  528$\pm$48  &  493$\pm$42  &  470$\pm$39  &  783$\pm$57  & 1098$\pm$46 \\
\rowcolor{ours!15}
FB-CMA-ME-Iso      & 3603$\pm$44  & \best{667$\pm$58}  & \best{569$\pm$55}  & \best{641$\pm$56}  & \best{1040$\pm$43}  & \second{1208$\pm$40} \\
\rowcolor{ours!15}
FB-BI    & \best{4074$\pm$60}  &   70$\pm$48  &  510$\pm$52  &  163$\pm$44  &  457$\pm$47  & 1185$\pm$54 \\
\rowcolor{ours!15}
FB-PGA-ME       & \second{3896$\pm$66}  &   63$\pm$43  &  514$\pm$45  &  176$\pm$43  &  475$\pm$49  & \best{1278$\pm$57} \\
\bottomrule
\end{tabular}
}
\end{table}

\begin{table}[ht]
\centering
\small
\setlength{\tabcolsep}{6pt}
\caption{QD-scores on \textbf{Cube-Single} (mean\,$\pm$\,std over 5 seeds).
  \textbf{Bold}: best; \underline{underline}: second best.}
\label{tab:results_cube}
\begin{tabular}{lcccc}
\toprule
\textbf{Method}
  & \textbf{Energy Eff.\ $(x,y)$}
  & \textbf{Energy Eff.\ $(x,z)$}
  & \textbf{Wrist Config}
  & \textbf{Grasp Style} \\
\midrule
\multicolumn{5}{l}{\textit{Baselines}} \\[1pt]
\rowcolor{baselinebg}
MAP-Elites      &    0$\pm$0 &    2$\pm$0 &  305$\pm$0 &    6$\pm$0 \\
\rowcolor{baselinebg}
CMA-ME          &    4$\pm$0 &    2$\pm$0 &  443$\pm$0 &   29$\pm$0 \\
\rowcolor{baselinebg}
PGA-ME          &    0$\pm$0 &   89$\pm$0 &  680$\pm$5 &    8$\pm$1 \\
\rowcolor{baselinebg}
DCRL-ME         &    0$\pm$0 &  141$\pm$0 &  521$\pm$0 &    6$\pm$0 \\
\rowcolor{baselinebg}
ME+Pretrain     &    3$\pm$0 &    1$\pm$0 &  152$\pm$0 &    4$\pm$0 \\
\rowcolor{baselinebg}
DDE-Elites          &  875$\pm$4 &  174$\pm$0 &    0$\pm$0 &    0$\pm$0 \\
\addlinespace[3pt]
\multicolumn{5}{l}{\textit{BFM-QD (ours)}} \\[1pt]
\rowcolor{ours!15}
FB-ME           & 2486$\pm$83  & 4240$\pm$89  & 1783$\pm$105 &  229$\pm$103 \\
\rowcolor{ours!15}
FB-CMA-ME       & \second{5368$\pm$106} & \second{5095$\pm$93}  & \best{1953$\pm$120} & \second{1073$\pm$111} \\
\rowcolor{ours!15}
TD-JEPA-CMA-ME  & 5018$\pm$121 & 3888$\pm$106 & 1690$\pm$87  &  290$\pm$144 \\
\rowcolor{ours!15}
Lap-CMA-ME          &  613$\pm$73  &  407$\pm$93  &  320$\pm$114 &   17$\pm$100 \\
\rowcolor{ours!15}
BYOL$\gamma$-CMA-ME         &  657$\pm$77  & 1115$\pm$96  &  317$\pm$109 &   57$\pm$110 \\
\rowcolor{ours!15}
BTD-FB-CMA-ME   & 5192$\pm$107 & 4963$\pm$84  & 1906$\pm$113 &  919$\pm$110 \\
\rowcolor{ours!15}
FB-CMA-ME-Iso      & \best{5965$\pm$114} & \best{5957$\pm$93}  & \second{1945$\pm$121} & \best{1227$\pm$108} \\
\rowcolor{ours!15}
FB-BI    & 2787$\pm$80  & 4892$\pm$80  & 1928$\pm$97  &  255$\pm$93  \\
\rowcolor{ours!15}
FB-PGA-ME       & 2835$\pm$74  & 4832$\pm$73  & 1887$\pm$94  &  255$\pm$101 \\
\bottomrule
\end{tabular}
\end{table}
\newpage



\newpage
\section*{NeurIPS Paper Checklist}

\begin{enumerate}

\item {\bf Claims}
    \item[] Question: Do the main claims made in the abstract and introduction accurately reflect the paper's contributions and scope?
    \item[] Answer: \answerYes{} 
    \item[] Justification: The abstract and introduction accurately state all contributions. Claims are matched by the experiments in Section~\ref{section:experiments}.

\item {\bf Limitations}
    \item[] Question: Does the paper discuss the limitations of the work performed by the authors?
    \item[] Answer: \answerYes{} 
    \item[] Justification: Section~\ref{section:limitation} explicitly discusses three limitations: (i) requirement of a diverse offline dataset; (ii) performance degradation with narrower datasets; (iii) the frozen BFM imposes a hard ceiling on behavioral expressivity.

\item {\bf Theory assumptions and proofs}
    \item[] Question: For each theoretical result, does the paper provide the full set of assumptions and a complete (and correct) proof?
    \item[] Answer: \answerYes{} 
    \item[] Justification: See \ref{method:bi}, and a complete proof is provided in \ref{app:proposition}. 

    \item {\bf Experimental result reproducibility}
    \item[] Question: Does the paper fully disclose all the information needed to reproduce the main experimental results of the paper to the extent that it affects the main claims and/or conclusions of the paper (regardless of whether the code and data are provided or not)?
    \item[] Answer: \answerYes{} 
    \item[] Justification: Appendix~\ref{app:experimental_details} provides full experimental details: environments, datasets, archive configuration, evaluation budget, MLP architecture, latent dimension, all hyperparameters, number of seeds, and the variation operators. 
    

\item {\bf Open access to data and code}
    \item[] Question: Does the paper provide open access to the data and code, with sufficient instructions to faithfully reproduce the main experimental results, as described in supplemental material?
    \item[] Answer: \answerNo{} 
    \item[] Justification: All datasets used are publicly available, and Appendix~\ref{app:experimental_details} provides sufficient detail for reimplementation.

\item {\bf Experimental setting/details}
    \item[] Question: Does the paper specify all the training and test details (e.g., data splits, hyperparameters, how they were chosen, type of optimizer) necessary to understand the results?
    \item[] Answer: \answerYes{} 
    \item[] Justification: All training and evaluation details are reported in Appendix~\ref{app:experimental_details}.

\item {\bf Experiment statistical significance}
    \item[] Question: Does the paper report error bars suitably and correctly defined or other appropriate information about the statistical significance of the experiments?
    \item[] Answer: \answerYes{} 
    \item[] Justification: All results are averaged over 5 (3 in two additional experiments) independent random seeds. Shaded regions in learning curves and error bars in bar plots denote standard deviation across seeds.

\item {\bf Experiments compute resources}
    \item[] Question: For each experiment, does the paper provide sufficient information on the computer resources (type of compute workers, memory, time of execution) needed to reproduce the experiments?
    \item[] Answer: \answerYes{} 
    \item[] Justification: Compute resources mentioned in the Appendix~\ref{app:compute}
    
\item {\bf Code of ethics}
    \item[] Question: Does the research conducted in the paper conform, in every respect, with the NeurIPS Code of Ethics \url{https://neurips.cc/public/EthicsGuidelines}?
    \item[] Answer: \answerYes{} 
    \item[] Justification: The research involves only simulated benchmarks. No human subjects, sensitive data, or dual-use concerns are present

\item {\bf Broader impacts}
    \item[] Question: Does the paper discuss both potential positive societal impacts and negative societal impacts of the work performed?
    \item[] Answer: \answerNA{} 
    \item[] Justification: This is foundational research on reinforcement learning algorithms using simulated environments. No involvement of real deployments, sensitive data, or high-risk technology.
    
\item {\bf Safeguards}
    \item[] Question: Does the paper describe safeguards that have been put in place for responsible release of data or models that have a high risk for misuse (e.g., pre-trained language models, image generators, or scraped datasets)?
    \item[] Answer: \answerNA{} 
    \item[] Justification: All datasets used are existing public benchmarks. No safeguards are required.

\item {\bf Licenses for existing assets}
    \item[] Question: Are the creators or original owners of assets (e.g., code, data, models), used in the paper, properly credited and are the license and terms of use explicitly mentioned and properly respected?
    \item[] Answer: \answerYes{} 
    \item[] Justification: All assets (libraries and baselines) are properly cited.

\item {\bf New assets}
    \item[] Question: Are new assets introduced in the paper well documented and is the documentation provided alongside the assets?
    \item[] Answer: \answerNA{} 
    \item[] Justification: The contribution is a framework and algorithmic approach.

\item {\bf Crowdsourcing and research with human subjects}
    \item[] Question: For crowdsourcing experiments and research with human subjects, does the paper include the full text of instructions given to participants and screenshots, if applicable, as well as details about compensation (if any)? 
    \item[] Answer: \answerNA{} 
    \item[] Justification: The paper involves no crowdsourcing and no research with human subjects. All experiments are conducted entirely in simulation.

\item {\bf Institutional review board (IRB) approvals or equivalent for research with human subjects}
    \item[] Question: Does the paper describe potential risks incurred by study participants, whether such risks were disclosed to the subjects, and whether Institutional Review Board (IRB) approvals (or an equivalent approval/review based on the requirements of your country or institution) were obtained?
    \item[] Answer: \answerNA{} 
    \item[] Justification: No human subjects are involved. IRB approval is not required.

\item {\bf Declaration of LLM usage}
    \item[] Question: Does the paper describe the usage of LLMs if it is an important, original, or non-standard component of the core methods in this research? Note that if the LLM is used only for writing, editing, or formatting purposes and does \emph{not} impact the core methodology, scientific rigor, or originality of the research, declaration is not required.
    \item[] Answer: \answerNA{} 
    \item[] Justification: No LLMs needed in our research.

\end{enumerate}

\end{document}